\documentclass[journal]{IEEEtran}
\usepackage{graphicx}
\usepackage{subcaption}
\usepackage[table,xcdraw]{xcolor}
\usepackage{multirow}
\usepackage{adjustbox}
\usepackage{subfig}
\usepackage{booktabs}
\usepackage{cite}
\usepackage{bbding}
\usepackage{amsmath}
\usepackage{amssymb}
\usepackage{booktabs}
\usepackage{makecell}
\usepackage{multirow}
\usepackage{adjustbox}
\usepackage{booktabs}
\usepackage{multirow}
\usepackage{graphicx}

\definecolor{darkgreen}{RGB}{0,150,0}
\definecolor{others}{rgb}{0, 0, 0}
\definecolor{barrier}{rgb}{1, 0.47058824, 0.19607843}
\definecolor{bicycle}{rgb}{1, 0.75294118, 0.79607843}
\definecolor{bus}{rgb}{1, 1, 0.0}
\definecolor{car}{rgb}{0.0, 0.58823529, 0.96078431}
\definecolor{construction}{rgb}{0, 1, 1}
\definecolor{motorcycle}{rgb}{1, 0.49803922, 0}
\definecolor{pedestrian}{rgb}{1, 0, 0}
\definecolor{cone}{rgb}{1, 0.94117647, 0.58823529}
\definecolor{trailer}{rgb}{0.52941176, 0.23529412, 0}
\definecolor{truck}{rgb}{0.62745098, 0.1254902, 0.94117647}
\definecolor{driveable}{rgb}{1, 0, 1}
\definecolor{flat}{rgb}{0.54509804,0.5372549,0.5372549}
\definecolor{sidewalk}{rgb}{0.29411765,0,0.29411765}
\definecolor{terrain}{rgb}{0.58823529,0.94117647,0.31372549}
\definecolor{manmade}{rgb}{0.90196078,0.90196078,0.98039216}
\definecolor{vegetation}{rgb}{0,0.68627451,0}
\definecolor{ego_vehicle}{rgb}{0,0,0}
\definecolor{scar}{rgb}{0.39215686, 0.58823529, 0.96078431}
\definecolor{sbicycle}{rgb}{0.39215686, 0.90196078, 0.96078431}
\definecolor{smotorcycle}{rgb}{0.11764706, 0.23529412, 0.58823529}
\definecolor{struck}{rgb}{0.31372549, 0.11764706, 0.70588235}
\definecolor{sother-vehicle}{rgb}{0.39215686, 0.31372549, 0.98039216}
\definecolor{sperson}{rgb}{1.        , 0.11764706, 0.11764706}
\definecolor{sbicyclist}{rgb}{1.        , 0.15686275, 0.78431373}
\definecolor{smotorcyclist}{rgb}{0.58823529, 0.11764706, 0.35294118}
\definecolor{sroad}{rgb}{1.        , 0.        , 1.        }
\definecolor{sparking}{rgb}{1.        , 0.58823529, 1.        }
\definecolor{ssidewalk}{rgb}{0.29411765, 0.        , 0.29411765}
\definecolor{sother-ground}{rgb}{0.68627451, 0.        , 0.29411765}
\definecolor{sbuilding}{rgb}{1.        , 0.78431373, 0.        }
\definecolor{sfence}{rgb}{1.        , 0.47058824, 0.19607843}
\definecolor{svegetation}{rgb}{0.        , 0.68627451, 0.        }
\definecolor{strunk}{rgb}{0.52941176, 0.23529412, 0.        }
\definecolor{sterrain}{rgb}{0.58823529, 0.94117647, 0.31372549}
\definecolor{spole}{rgb}{1.        , 0.94117647, 0.58823529}
\definecolor{straffic-sign}{rgb}{1.        , 0.        , 0.        }
\ifCLASSINFOpdf

\else
 
\fi

\begin{document}

\title{TFusionOcc: T-Primitive Based Object-Centric Multi-Sensor Fusion Framework for 3D Occupancy Prediction}

\author{Zhenxing~Ming$^1$, Yaoqi~Huang$^1$, Julie~Stephany~Berrio$^2$, Mao~Shan$^1$, and Stewart~Worrall$^1$
\thanks{$^1$ The authors are with the Australian Centre for Robotics (ACFR) at the University of Sydney (NSW, Australia). E-mails: zmin2675@uni.sydney.edu.au, \{yoki.huang, mao.shan, stewart.worrall\}@sydney.edu.au}%
\thanks{$^2$ The author is with the School of Mechanical and Mining Engineering at the University of Queensland (QLD, Australia). E-mail: s.berrioperez@uq.edu.au}%

}
\maketitle

\begin{abstract}
The prediction of 3D semantic occupancy enables autonomous vehicles (AVs) to perceive the fine-grained geometric and semantic scene structure for safe navigation and decision-making. Existing methods mainly rely on either voxel-based representations, which incur redundant computation over empty regions, or object-centric Gaussian primitives, whose ellipsoidal per-primitive geometry can limit representation efficiency for complex, non-convex, and asymmetric structures under a fixed primitive and computational budget.
In this paper, we present TFusionOcc, a T-primitive-based object-centric multi-sensor fusion framework for 3D semantic occupancy prediction. Specifically, we introduce a family of Student’s t-distribution-based T-primitives, including the plain T-primitive, T-Superquadric, and deformable T-Superquadric with inverse warping, where the deformable T-Superquadric serves as the key geometry-enhancing primitive. We further develop a unified probabilistic formulation based on the Student’s t-distribution and the T-mixture model (TMM) to jointly model occupancy and semantics, and design a tightly coupled multi-stage fusion architecture to integrate camera and LiDAR cues effectively. Extensive experiments on nuScenes demonstrate competitive performance against recent state-of-the-art methods and strong performance among object-centric multi-sensor approaches, while additional evaluations on nuScenes-C show robust performance across most corruption scenarios. Further evaluation on the separate SemanticKITTI benchmark demonstrates cross-dataset applicability. The code will be available at: https://github.com/DanielMing123/TFusionOcc 
\end{abstract}


\IEEEpeerreviewmaketitle

\section{Introduction}
\IEEEPARstart{A}{ccurate} and efficient scene representation is essential for autonomous driving, since downstream planning and decision-making rely on a reliable understanding of both geometry and semantics. Compared with conventional 3D object detection and BEV segmentation, 3D semantic occupancy prediction provides a more complete and denser description of the surrounding environment, especially for irregular structures, free space, and partially occluded regions. Recent progress in vision-centric 3D semantic occupancy prediction methods \cite{gaussianformer, semantic, rethinking, navigscene, 3dvislan} has explored improvements in scene representation, 2D-to-3D transformation, and temporal modeling. Despite these advances, vision-centric approaches still rely on indirectly inferred 3D geometry from image observations, motivating the integration of complementary geometric measurements from additional sensors. Multi-sensor fusion further improves this representation by combining the complementary strengths of cameras and LiDAR.
\begin{figure}[t]
     \centering
     \begin{subfigure}[]{\columnwidth}
         \centering
        \includegraphics[width=\columnwidth]{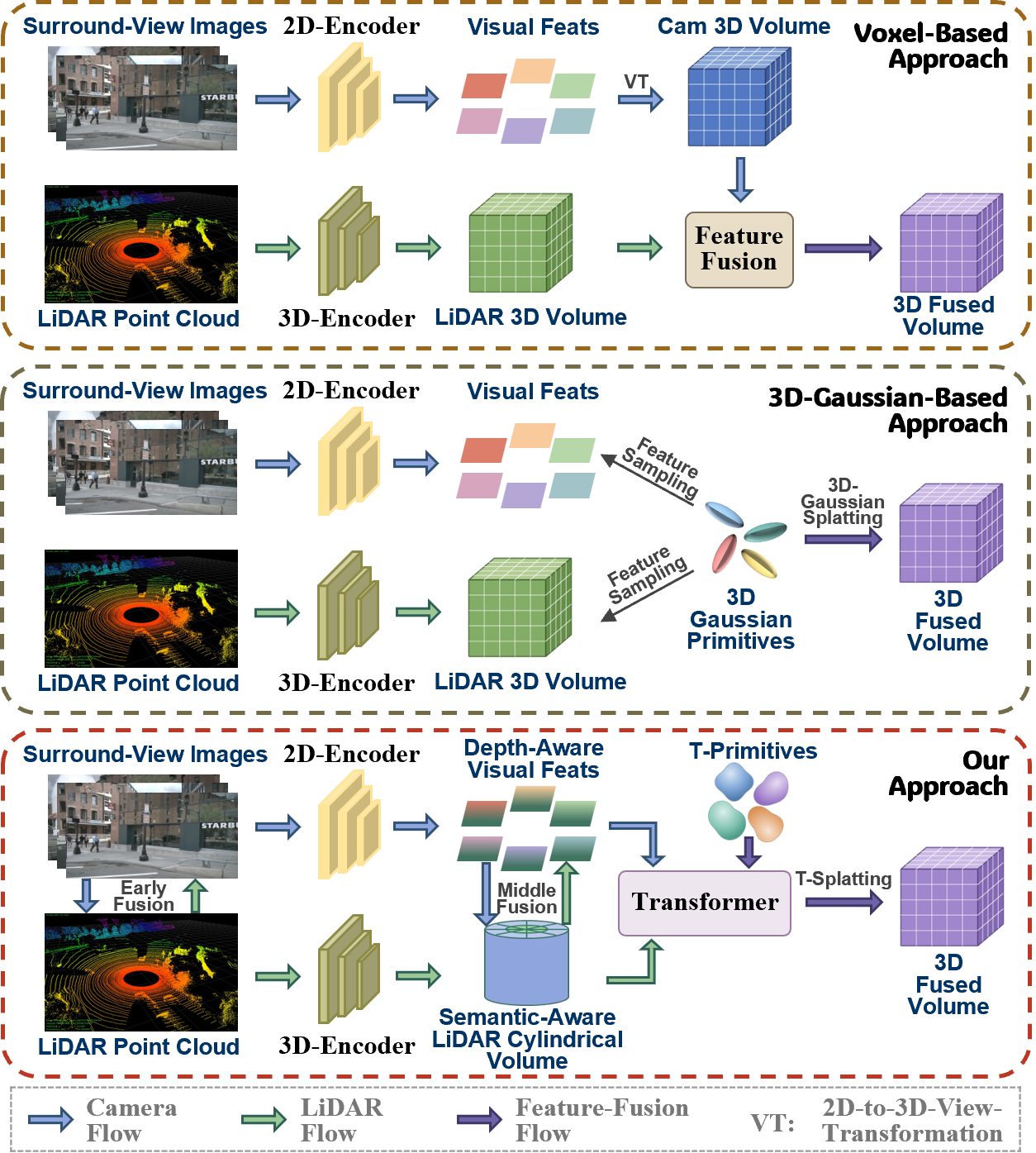}
     \end{subfigure}
        \caption{\small \textbf{Pipeline of three approaches:} Voxel-based approach (top), 3D-Gaussian-Primitive-based object-centric approach (middle), and our approach (bottom).}
        \label{teaser}
\end{figure}

Recent progress in multi-sensor 3D semantic occupancy prediction has been largely driven by voxel-based fusion methods \cite{occfusion,co-occ,fusionocc,occcylindrical,daocc,sdgocc,effocc}, which integrate complementary signals from cameras and LiDAR. Although these methods have achieved strong performance, they generally rely on dense volumetric processing (Fig. \ref{teaser}, voxel-based approach), leading to substantial redundant computation over empty regions that dominate outdoor driving scenes. This inefficiency has motivated the development of object-centric representations, where a compact set of learnable primitives is used to model occupied regions more selectively. Among object-centric approaches \cite{gaussianformer3d,gaussianfusionocc}, Gaussian-based primitives have recently shown promise for improving efficiency and compact scene representation (Fig. \ref{teaser}, 3D Gaussian-based approach). However, the ellipsoidal geometry of individual Gaussians may require multiple primitives to approximate non-convex, asymmetric, and structurally irregular scene elements, reducing representation efficiency under a fixed primitive and computational budget. Superquadrics \cite{quadricformer} offer greater per-primitive geometric flexibility via learnable shape exponents, but conventional superquadrics remain constrained by their largely symmetric geometry. These limitations motivate the development of more expressive deformable primitives that can represent complex scene structures more efficiently under a fixed primitive budget. In addition, Gaussian-like formulations are less effective in handling outliers and long-tailed spatial distributions, which may reduce robustness under noisy or corrupted observations. As a result, existing object-centric occupancy frameworks still face a performance gap when compared with strong voxel-based methods, especially when an accurate geometric representation is required.

To address these limitations, we propose TFusionOcc (Fig \ref{teaser}, our approach), a T-primitive-based object-centric multi-sensor fusion framework for 3D semantic occupancy prediction. Specifically, we introduce a family of Student’s t-distribution-based T-primitives, including the plain T-primitive, T-Superquadric and a deformable T-Superquadric with inverse warping, where the latter serves as the key geometry-enhancing primitive. We further develop a unified probabilistic formulation based on Student's t-distribution and the T-mixture model (TMM) to jointly model occupancy and semantics, and design a compact multi-stage fusion pipeline to effectively integrate camera and LiDAR cues. In this way, TFusionOcc improves geometric representation while preserving the efficiency advantage of object-centric modeling. Extensive experiments on nuScenes \cite{nuscenes} show that TFusionOcc achieves competitive performance against recent state-of-the-art methods while demonstrating strong performance among object-centric multi-sensor approaches and consistently benefiting from stronger primitive geometry. In particular, our results indicate that the flexibility of primitive shapes is more decisive than simply increasing the primitive count, highlighting the importance of geometry-aware primitive design. To further evaluate the generalizability of the proposed framework beyond nuScenes, we conduct experiments on the separate SemanticKITTI dataset \cite{semantickitti}. TFusionOcc achieves 24.51\% mIoU on the SemanticKITTI validation set, and the consistent performance improvement from T-P to T-SQ and T-SQ-IW further indicates that the benefit of stronger primitive geometry transfers across substantially different dataset domains and sensor configurations. Additional evaluations on nuScenes-C \cite{nuscenes-c} in different corruption settings further demonstrate strong robustness in many challenging cases.  The main contributions of this paper are summarized as follows: 
\begin{itemize}
    \item We propose a family of Student’s t-distribution-based T-primitives for object-centric 3D occupancy prediction, including the plain T-primitive, T-Superquadric, and a deformable T-Superquadric with inverse warping. In particular, the deformable T-Superquadric significantly improves geometric modeling flexibility, enabling better representation of non-convex and asymmetric structures. 
    \item We develop a unified probabilistic framework based on a Student’s t-shaped heavy-tailed kernel and T-mixture formulation for joint occupancy and semantic modeling. Compared with the Gaussian kernel, it improves robustness for handling outliers and long-tailed spatial responses and consistently enhances occupancy prediction. Together with more expressive primitive geometry, both probabilistic modeling and geometric flexibility contribute complementary performance gains. 
    \item We design a tightly coupled multi-sensor fusion pipeline to support effective T-primitive learning. Specifically, it includes fused-depth-guided visual feature lifting, transformer-based T-primitive-centric refinement, and the GCWSFusion module for adaptive cross-modal interaction, allowing T-primitives to better exploit cross-modal cues for 3D semantic occupancy prediction.
\end{itemize}

The remainder of this paper is structured as follows. Section \ref{literature} provides an overview of related research and identifies the key differences between this study and previous publications. Section \ref{model} outlines the general framework of TFusionOcc and offers a detailed explanation of the implementation of each module. Section \ref{simulation} presents the results of our experiments. Finally, Section \ref{conclusion} provides the conclusion of our work.

\section{Related Work}\label{literature}
\subsection{Voxel-Based Approach}
In OccFusion \cite{occfusion}, the authors extract a set of 3D voxel volumes from each different modality sensor, such as surround-view cameras, surround-view radars and LiDAR, followed by the feature fusion operation through the proposed dynamic feature fusion module to perform multi-sensor fusion-based 3D semantic occupancy prediction. The model exhibits strong robustness under challenging rain and night scenarios. In Co-Occ \cite{co-occ}, the authors leverage the volume rendering technique, which was originally derived from NeRF \cite{nerf}, to refine the intermediate 3D feature volumes. They also proposed a GSFusion module for efficient multimodal feature fusion, achieving SOTA performance. In FusionOcc \cite{fusionocc}, the authors proposed a cross-modality fusion module to generate a high-quality depth-aware visual feature, followed by a 2D to 3D view transformation, resulting in volumes of depth-aware visual features under the Cartesian coordinate system. The LiDAR branch conducts voxelization and voxel encoding, resulting in a LiDAR feature volume also under the Cartesian coordinate system. Different modality feature volumes were fused via feature channel concatenation and further refined by a 3D Encoder. In OccCylindrical \cite{occcylindrical}, the authors proposed a multi-sensor fusion-based framework that uses a cylindrical partition to perform voxelisation of a LiDAR-captured 3D point cloud and a pseudo 3D point cloud predicted by surround-view cameras, resulting in better preservation of 3D geometric information. The feature fusion and further refinement operations are performed under the cylindrical coordinate system. In AdaptiveOcc \cite{adaptiveocc}, the author explored using an efficient octree data structure to preserve the fine-grained geometry of the 3D scene, achieving a good balance between computational complexity and prediction performance for the voxel-grid-based approach. In DAOcc \cite{daocc}, the authors leverage multi-task learning to introduce an additional 3D supervision signal to guide intermediate features, leading to robust generalization and overall SOTA performance. Similarly, in Doracamom \cite{doracamom}, the authors developed a joint 3D detection and occupancy prediction pipeline under a camera-radar fusion paradigm. By leveraging the strengths of multi-task learning and further incorporating temporal fusion, the proposed method achieved superior performance in both 3D object detection and 3D semantic occupancy prediction. In addition, SDGOcc \cite{sdgocc} incorporates a multi-task head to introduce a 2D semantic mask into the model training procedure, leveraging an additional 2D semantic mask to enhance the precision and densification of depth maps obtained from LiDAR point cloud projection, thereby boosting the model's overall performance.

Despite achieving remarkable performance, these voxel-based methods incur redundant calculations for empty voxels, which are often dominant in 3D driving scenarios, resulting in inefficient computation. Our study presents a novel object-centric approach, aiming to address the aforementioned constraints by leveraging T-Primitives to model only occupied space in driving scenarios.

\subsection{Object-Centric Modeling Approach}
In the GaussianFormer series \cite{gaussianformer,gaussianformer2}, the authors proposed an object-centric 3D Gaussian model together with a Gaussian mixture model (GMM) to improve computational efficiency and robustness. In GraphGSOcc \cite{graphgsocc}, the authors combine the object-centric 3D Gaussian representation with a graph transformer to realize dynamic and static object decoupling modeling. Allow the model to optimize foreground dynamic objects and background static objects separately. The approach further incorporates a temporal fusion mechanism to achieve superior performance. In GaussianFormer3D \cite{gaussianformer3d}, the authors extend the GaussianFormer work to a multi-sensor fusion scenario. Using a set of 3D Gaussian primitives as an intermediate feature representation, the authors designed a pipeline that enables each 3D Gaussian primitive to aggregate features from different modalities and fuse them to perform 3D primitive to-3D voxel volume Gaussian splatting. Benefiting from the object-centric modeling technique and multi-modal fusion, the model achieves significant performance and efficiency improvements. In GaussianFusionOcc \cite{gaussianfusionocc}, the authors proposed a GaussianFusionBlock module that performs efficient and high-quality feature aggregation and fusion across different modalities, allowing high-quality 3D Gaussian primitive modeling and leading to outstanding performance.

Unlike previous object-centric methods, our work introduces a family of Student’s t-distribution-based T-primitives for 3D semantic occupancy prediction. In particular, we propose a deformable T-Superquadric with inverse warping, which substantially strengthens primitive-level geometric flexibility. We further formulate occupancy and semantics under a unified TMM framework and design a multi-stage camera–LiDAR fusion pipeline to support effective T-primitive refinement. In this way, our method aims to preserve the efficiency advantage of object-centric modeling while improving geometric representation capability and robustness.

\section{TFusionOcc}\label{model}
This work leverages a set of Student-T-distribution-based primitives combined with a 3D to 3D splatting operation to generate a dense 3D semantic occupancy grid of the surrounding scene by integrating information from surround-view cameras and LiDAR. Thus, the problem can be formulated as follows:
\begin{equation}
    P^{T}_{refined}=F\left ( Cam^{1},Cam^{2},...,Cam^{N},Lidar,P^{T} \right )
\end{equation}
\begin{equation}
    Occ = Splat\left ( P^{T}_{refined} \right ) 
\end{equation}
where $F$ refers to a multi-sensor fusion framework, $P^{T}$ refers to a set of primitives based on T-distribution $P^{T}=\left \{ p^{T}_{1},p^{T}_{2},...,p^{T}_{K} \right \}$, along with a set of query vectors $Q = \left \{ q_{1},q_{2},...,q_{K} \right \}$, $P^{T}_{refined}$ refers to a set of T-distribution-based primitives refined through multi-sensor fusion framework, $N$ refers to the total number of surround-view images and $Splat\left(\cdot\right)$ refers to the splatting operation. In the final predicted 3D semantic occupancy $Occ\in R^{\left \{ X\times Y\times Z \right \}} $, each grid is assigned a semantic property ranging from 0 to $C$, where $C$ refers to the total number of semantic classes. In our case, a class value of 0 corresponds to an empty grid.
\subsection{Overview}
\begin{figure*}[htbp]
     \centering
        \includegraphics[width=\textwidth]{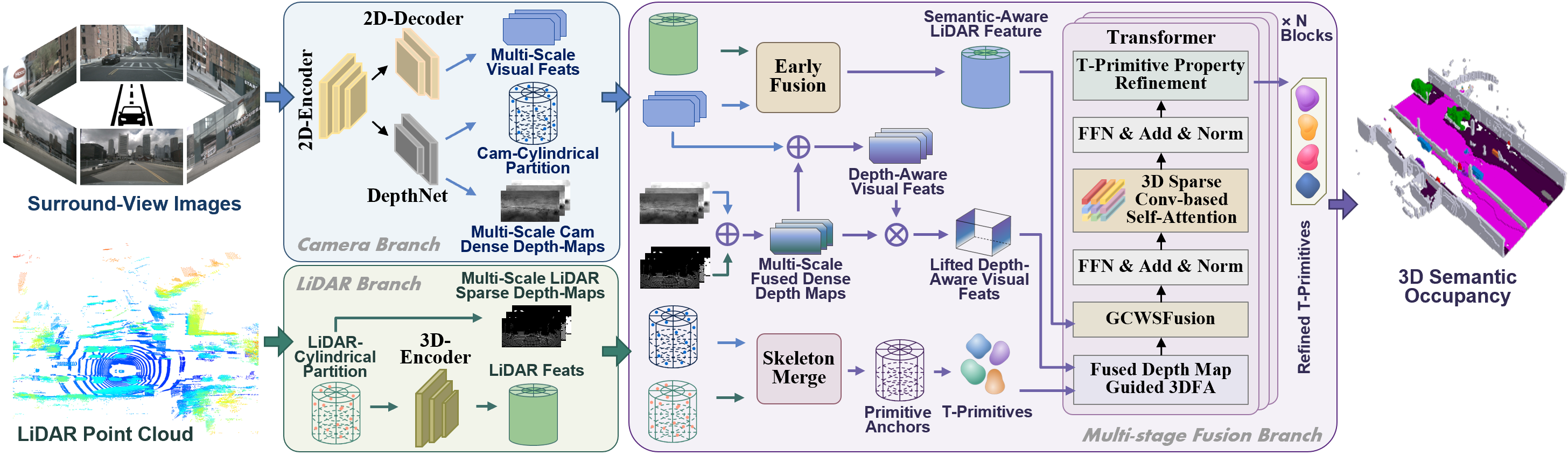}
        \caption{\small \textbf{Overall architecture of TFusionOcc.} The pipeline comprises a camera branch, a LiDAR branch, and a multi-stage fusion branch. The camera branch extracts multi-scale visual features and predicts a pseudo 3D point cloud from surround-view images. The pseudo 3D point cloud is further projected and cylindrically partitioned, resulting in camera-based, multi-scale, dense depth maps and a voxel volume defined under cylindrical coordinates. The LiDAR branch applies a cylindrical partition followed by a 3D encoder to extract the LiDAR feature. Meanwhile, the LiDAR point cloud is also projected to generate LiDAR-based, multi-scale, sparse depth maps. The feature fusion branch adopts a multi-stage fusion strategy to merge all outputs from the two-modality branch and leverages a proposed transformer to refine the T-primitives property through fused features.}
        \label{TFusionOcc}
\end{figure*}
The overall architecture is shown in Fig.~\ref{TFusionOcc}. For the camera branch, given $\left \{ img^{1},img^{2},...,img^{N} \right \}$ surround-view images, a 2D encoder and a decoder are leveraged to extract $V^{Cam}=\left \{ \left \{ V_{n}^{l} \right \}_{n=1}^{N}\in \mathbb{R}^{C \times H_{l} \times W_{l}} \right \} _{l=1}^{L=3}$ multi-scale visual features. Meanwhile, a DepthNet is leveraged, resulting in a camera-based pseudo 3D point cloud $Pt^{cam}$ and a multi-scale camera-based dense depth maps  $V_{depth}^{cam}=\left \{ \left \{ V_{depth\_n}^{l} \right \}_{n=1}^{N}\in \mathbb{R}^{D \times H_{l} \times W_{l}} \right \} _{l=1}^{L=3}$, where $D$ refers to the total amount of depth-bins. The pseudo 3D point cloud $Pt^{cam}$ is further divided into cylindrical voxels, resulting in a camera-based 3D voxel volume $Vol^{Cam}_{Cylin}$ defined in cylindrical coordinates. For the LiDAR branch, given a 3D point cloud $Pt^{lidar}$ from LiDAR, the cylindrical partition is first applied, resulting in a LiDAR-based 3D voxel volume $Vol^{Lidar}_{Cylin}$. Then a 3D encoder (e.g., Cylinder3D \cite{cylinder3d}) is used to extract LiDAR features $V^{Lidar}$ in cylindrical coordinates. Meanwhile, $Pt^{Lidar}$ is projected onto each surround-view camera frame, resulting in sparse multi-scale LiDAR-based depth maps $V_{depth}^{lidar}=\left \{ \left \{ V_{depth\_n}^{l} \right \}_{n=1}^{N}\in \mathbb{R}^{D \times H_{l} \times W_{l}} \right \} _{l=1}^{L=3}$. Then, a multi-stage feature fusion strategy is adopted. For the camera branch side, $V_{depth}^{cam} \oplus V_{depth}^{lidar}$ results in multi-scale fused dense depth maps $V_{depth}^{fuse}$. Then $V_{depth}^{fuse}$ is encoded by a multi-layer perceptron (MLP) and added to $V^{Cam}$, resulting in multi-scale depth-aware visual features $F_{depth}^{Cam}$. In addition, an outer product operation $F_{depth}^{Cam}\otimes V_{depth}^{fuse}$ is conducted, resulting in a multi-scale lifted depth-aware visual feature $Vol^{Cam}_{depth}$. For the LiDAR branch side, each occupied voxel center of $Vol^{Lidar}_{Cylin}$ serves as an anchor to project onto $V^{Cam}$, thereby aggregating semantic information. The aggregated semantic information is further fused with the LiDAR feature $Vol^{Lidar}_{Cylin}$ through an early-fusion module, resulting in a semantic-aware LiDAR feature $Vol^{Lidar}_{sem}$. The $Vol^{Cam}_{Cylin}$ and $Vol^{Lidar}_{Cylin}$ are fused through the skeleton merge module, resulting in a fused voxel volume $Vol^{Fused}_{Cylin}$ under cylindrical coordinates. The $Vol^{Fused}_{Cylin}$ is then used as primitive anchors to initialize T-Primitives $P^{T}$ under cylindrical coordinates. The $P^{T}$, $Vol^{Cam}_{depth}$ and $Vol^{Lidar}_{sem}$ are fed to the transformer module, which stacks the $N\times$ blocks to obtain a set of refined T-Primitives. Lastly, the refined set of T-Primitives performs a splatting operation, resulting in the final 3D semantic occupancy grid. 

\subsection{Camera-based Pseudo 3D Point Cloud and Multi-Scale Cam Dense Depth Maps Generation}
\begin{figure}[htbp]
     \centering
        \includegraphics[width=\columnwidth]{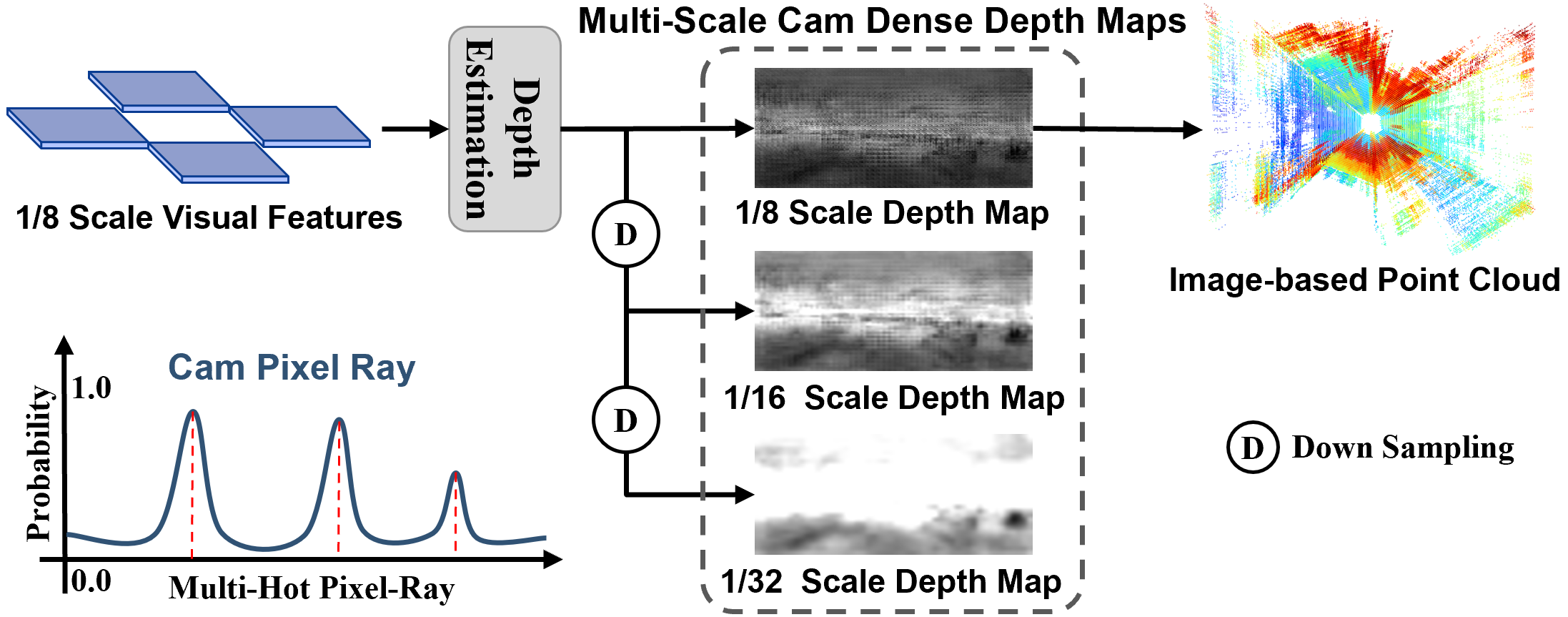}
        \caption{\small \textbf{Inner Structure of DepthNet.} The 1/8-scale visual features are first used to generate a 1/8-scale depth map. Then, bilinear interpolation is leveraged to generate 1/16- and 1/32-scale depth maps. Meanwhile, the image-based pseudo-point cloud is generated solely from a 1/8-scale depth map.}
        \label{DepthNet}
\end{figure}
The pseudo 3D point cloud $Pt^{cam}$ and camera-based multi-scale dense depth maps $V^{cam}_{depth}$ are generated through the DepthNet module as shown in Fig.~\ref{DepthNet}. The DepthNet module takes $V^{Cam}_{Encode}$ as input and leverages a SECONDFPN \cite{second} to generate a single-scale visual feature $V^{cam}_{sec}= \left \{ V_{n} \right \}_{n=1}^{N}\in \mathbb{R}^{C \times H_{l} \times W_{l}}$, whose resolution is $\frac{1}{8}$ of the input image. The $V^{cam}_{sec}$ is further passed to an MLP to perform the multi-hot per-ray depth estimation, resulting in a depth map $V_{depth\_1}\in \mathbb{R}^{N\times D \times H_{l} \times W_{l}}$ with each pixel ray having a set of depth bins separated by 0.5m intervals, and each depth bin on the pixel ray has a probability range in $[0, 1]$. Following \cite{gaussianformer2}, we adopt the 3D semantic occupancy label to supervise $V_{depth\_1}$. Then, bilinear interpolation with a 0.5 downsampling factor is applied iteratively to $V_{depth\_1}$, resulting in two additional depth maps that have $\frac{1}{16} $ and $\frac{1}{32} $ resolution of the input image. The predicted depth map and the two extra downsampled depth maps form $V_{depth}^{cam}=\left \{ \left \{ V_{depth\_n}^{l} \right \}_{n=1}^{N}\in \mathbb{R}^{D \times H_{l} \times W_{l}} \right \} _{l=1}^{L=3}$. Meanwhile, the pseudo 3D point cloud $Pt^{cam}$ is acquired only from $\frac{1}{8} $ scale depth map $V_{depth\_1}$.

\subsection{Multi-scale-Fused Dense Depth-Maps Generation}
\begin{figure}[htbp]
     \centering
         \includegraphics[width=\columnwidth]{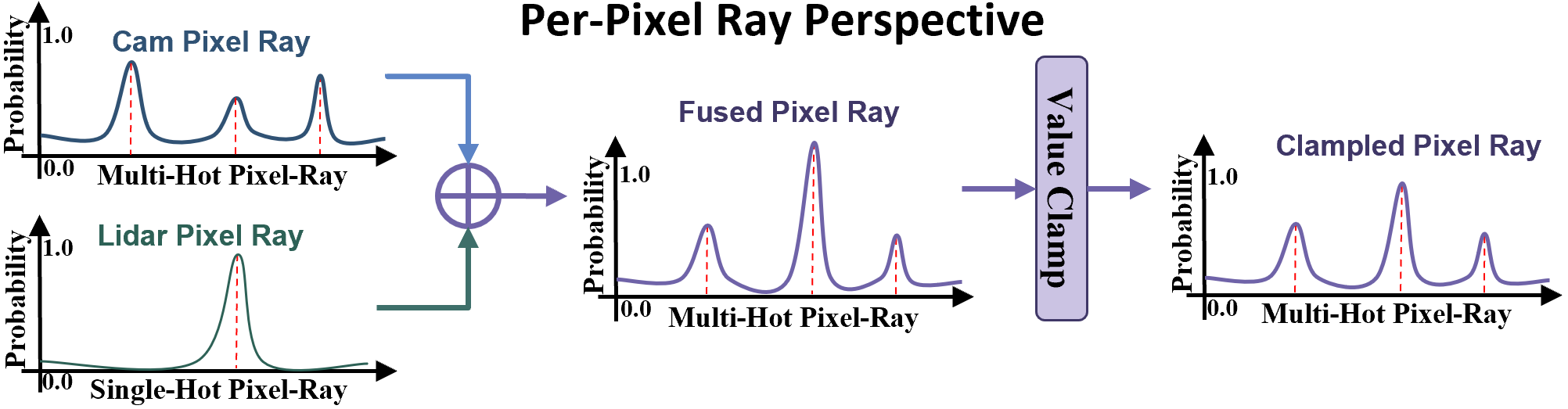}
        \caption{\small \textbf{Multi-Scale Fused Dense Depth Maps Generation.} This demonstrates the detailed multi-source depth map fusion from a per-pixel ray perspective.}
        \label{depth}
\end{figure}
Given the multi-scale dense depth maps from cameras, $V^{cam}_{depth}$, and the multi-scale sparse depth maps from LiDAR, $V^{lidar}_{depth}$, we perform element-wise addition at the per-pixel level, followed by a value clamping operation. This process produces the final multi-scale fused dense depth maps, as illustrated in Fig.~\ref{depth}. All depth maps from different sensors share a consistent representation, with the same number of depth bins $D$ and a uniform interval of 0.5 m between adjacent bins along each pixel ray. However, their value distributions differ: on the camera side, $V^{cam}_{depth}$ is represented as multi-hot per-pixel ray depth maps, where each depth bin takes a value within the range $[0, 1]$. In contrast, $V^{lidar}_{depth}$ is represented as single-hot per-pixel ray depth maps, where only one depth bin is assigned a value of 1.0. When the camera and LiDAR depth maps are combined via per-pixel addition, the resulting fused pixel rays may contain depth-bin values exceeding 1.0. To ensure valid probability values, all depth-bin responses are subsequently clamped to the range $[0, 1]$.

This fusion strategy combines the complementary strengths of the two modalities. The dense camera depth maps provide rich structural coverage but are less accurate, whereas the LiDAR depth maps are sparse but geometrically precise. Their combination yields denser, more reliable fused depth maps, which are subsequently used to enhance visual features and facilitate cross-modal fusion.

\subsection{Skeleton Merge Module}
\begin{figure}[htbp]
     \centering
         \includegraphics[width=\columnwidth]{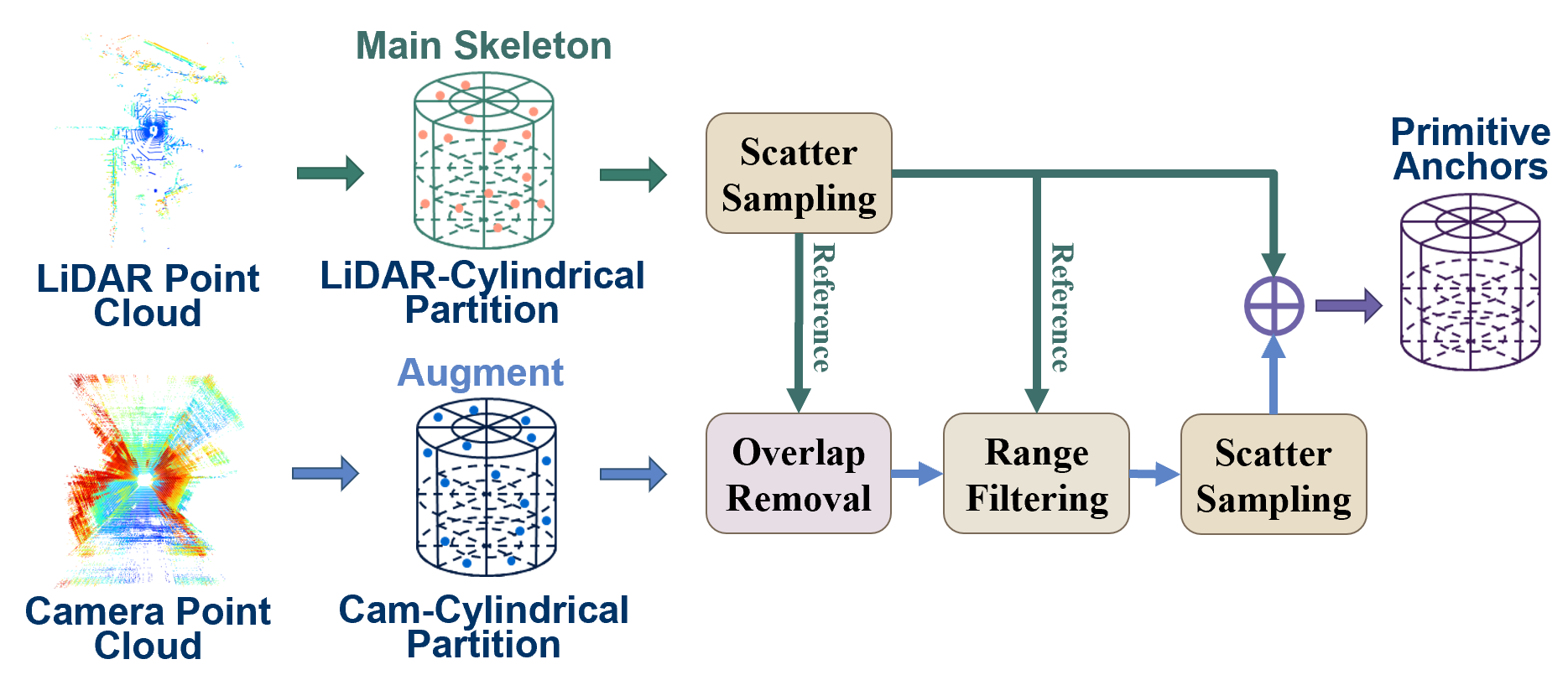}
        \caption{\small \textbf{Skeleton Merge Module.} The upper LiDAR-branch serves as the main skeleton to provide a foundation structure for the 3D scene, and the bottom camera-branch serves as an augmentation based on the main skeleton to provide more detailed local structure to compensate for the fine-grained geometry of the main skeleton.}
        \label{skeleton}
\end{figure}
The primitive anchor initialization process is shown in Fig.~\ref{skeleton}. 
Given the pseudo 3D point cloud derived from the camera $Pt^{cam}$ and the real 3D point cloud of the LiDAR $Pt^{lidar}$, we first apply the cylindrical partition to both point clouds, resulting in two volumes of voxels $Vol^{Cam}_{Cylin}$ and $Vol^{Lidar}_{Cylin}$ under cylindrical coordinates. We then predefine the total primitive anchors of $M+N$, where the anchors $M$ are sampled from $Vol^{Lidar}_{Cylin}$ and the anchors $N$ are sampled from $Vol^{Cam}_{Cylin}$, with a ratio of $M: N = 3:1$. We adopt the farthest point sampling strategy to obtain $M$ LiDAR anchors from $Vol^{Lidar}_{Cylin}$, which act as the main structural skeleton. Using these LiDAR anchors as reference, any voxel in $Vol^{Cam}_{Cylin}$ that spatially overlaps with the LiDAR anchors is removed. In addition, we impose a radius constraint $r=5$m to perform range filtering. Any voxel in $Vol^{Cam}_{Cylin}$ whose closest LiDAR anchor is farther away than $r$ is also discarded. Then, $N$ Camera anchors are obtained using the same sampling strategy on the remaining voxels in $Vol^{Cam}_{Cylin}$. Based on the aforementioned filtering and sampling strategy, we acquire a final set of $M+N$ anchors whose locations will be used to initialize a set of T-primitives. 

This design exploits the complementary strengths of the two modalities: LiDAR provides the main geometric structure with higher spatial accuracy, while the camera branch supplements local details and occluded regions. The resulting fused skeleton, therefore, offers a better initialization for subsequent T-primitive modeling.

\subsection{Early-Fusion Module}
\begin{figure}[htbp]
     \centering
        \includegraphics[width=\columnwidth]{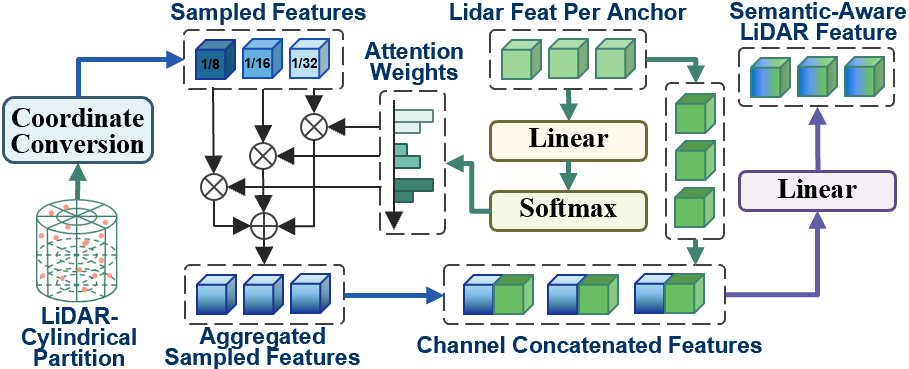}
        \caption{\small \textbf{Early-Fusion Module.} Each occupied voxel center in $Vol^{Lidar}_{Cylin}$ serves as an anchor and is projected onto multi-scale visual features to aggregate semantic information, yielding a semantic-aware LiDAR feature.}
        \label{earlyfusion}
\end{figure}
The detailed inner structure of the Early-Fusion module is shown in Fig.~\ref{earlyfusion}. 
Given the sparse volume of the voxel $Vol_{Cylin}^{Lidar}$ in cylindrical coordinates, we leverage each occupied voxel center as an anchor and convert each anchor position from cylindrical coordinates to Cartesian coordinates. Then, each anchor is projected onto multi-scale visual features $V^{Cam}$ to aggregate semantic information at three scales. The aggregated three-scale visual features are weighted, summed, and concatenated back to the corresponding voxel-associated LiDAR feature, then fed through an MLP layer to perform the feature fusion, resulting in the last semantic-aware LiDAR feature $Vol_{sem}^{Lidar}$.

This module injects image semantics into sparse LiDAR features at an early stage, allowing the LiDAR branch to retain its geometric precision while gaining richer semantic context from the camera branch.

\subsection{T-Primitive Representation}
We propose a probabilistic superposition of Student’s t-distribution-based primitives as an efficient and effective 3D scene representation. Specifically, we propose three types of primitives that are enhanced by the inner kernel of the T-distribution or T-distribution-like. The first is the Student-T-distribution-based plain primitive, which has the attributes as follows:
\begin{equation}
    P_{i}^{T}(m_{i},s_{i},r_{i},\alpha_{i},c_{i})   
\end{equation}
where $m_{i}$, $s_{i}$, $r_{i}$, $\alpha_{i}$ and $c_{i}$ refer to the mean, scale value for the $x,y,z$ axes, quaternion vector, opacity and semantic vector of the $i$-th primitive, respectively.
The second is the Student-T-distribution-based superquadric primitive, which has the attributes as follows:
\begin{equation}
    P_{SQi}^{T}(m_{i},s_{i},r_{i},\alpha_{i},\epsilon_{1_i},\epsilon_{2_i},c_{i})   
\end{equation}
where $\epsilon_{1}$ and $\epsilon_{2}$ are the shape exponents of the superquadric. The third is the Student-T-distribution-based superquadric with inverse-warp primitive that has the attributes as follows:
\begin{equation}
    P_{SQIWi}^{T}(m_{i},s_{i},r_{i},\alpha_{i},\epsilon_{1_i},\epsilon_{2_i},\left \{ \omega_{i,k} \right \} _{k=1}^{24},c_{i}) 
\end{equation}
where $\left \{ \omega_{i,k} \right \} _{k=1}^{24}$ is a set of learnable weighting scalers used to determine the magnitude of applied deformation, and each  $\omega_{i,k}$ has a value range between -1 and 1. 

Inspired by \cite{gaussianformer2}, we perform 3D geometry prediction and semantic prediction separately and adopt the multiplication theorem of probability and the T-mixture model to address them probabilistically.

\subsubsection{Student-T-Distribution and Multiplication Theorem Based Geometry Prediction}
In practice, we interpret the T-distribution primitives $P^{T}=\left \{ P^{T}_{i}  \right \}_{i=1}^{K} $ as the probability of their surrounding space being occupied. In detail, we assign a probability value of 100\% to the centers of the $P^{T}_{i}$, which decays with respect to the distance from the centers $m_{i}$. Assuming conditional independence of occupancy among different T-primitives, the final occupancy probability at x can be computed as:
\begin{equation}
    p_{o}\left ( x \right ) =1-\prod_{i=1}^{K}\left ( 1-p\left(x|P^{T}_{i}\right )\right) 
    \label{geo_pred}
\end{equation}
and
\begin{equation}
    p\left(x|P^{T}_{i}\right )=\left ( 1+\frac{d_{i}\left ( x \right ) }{\nu_i}  \right ) ^{-\frac{\nu_{i}+3}{2} } 
    \label{primi_prob}
\end{equation}
where $d_{i}\left ( x \right )\ge 0$ denotes the corresponding primitive-dependent geometric measure, and $\nu_{i}$ denotes the degrees of freedom of the $i$-th T-primitive. 
In our implementation, each primitive is associated with a rasterization radius \( \rho_i= \max\left\{ 1,\; \left\lceil \frac{\max(S_{xi},S_{yi},S_{zi})}{\Delta} \right\rceil \right\}, \) where \(\mathbf{s}_i=(S_{xi},S_{yi},S_{zi})\) contains the primitive’s scale parameters, \(\Delta\) is the voxel grid size.
Every T-primitive center lies inside the final voxel volume grid. $N_{i}=(2\cdot \rho_i)^{3}$ is the cardinality of the clipped, axis-aligned candidate window used to rasterize the \(i\)-th T-primitive. We further chose $\nu_{i}=N_{i}-1$ as a support-adaptive tail parameter—an inductive design choice rather than a statistically estimated degree of freedom. Hence, $N_{i} \ge 8$ and $\nu_{i} \ge 7$ ensure the existence of the mean and covariance of the corresponding Student’s t distribution. The discrete quantities $N_{i}$ and $\nu_{i}$ are intentionally treated as stop-gradient rasterization metadata. They are recomputed during every forward pass but held fixed during backpropagation. Therefore, 
\begin{equation}
    \frac{\partial N_{i}}{\partial \theta_{i}}= \frac{\partial \nu_{i}}{\partial \theta_{i}}=0
\end{equation}
where $\theta_{i}$ denotes the parameters of the $i$-th T-primitive. Gradients still flow through these parameters within the current rasterization window, making the renderer piecewise differentiable. Moreover, each T-primitive is evaluated only within its finite rasterization window and contributes exactly zero outside it, preventing leakage into distant regions. Within this bounded support, a smaller \(N_i\) produces a smaller \(\nu_i\) and hence a heavier-tailed kernel profile, whereas a larger \(N_i\) produces a more concentrated
profile. This support-adaptive design allows each T-primitive to adjust
its kernel according to its rasterization extent. $p \left ( x| P^{T}_{i} \right )$ denotes the probability of point x being occupied induced by the $i$-th T-primitive, and it has range $[0\sim 1]$. Depending on the primitive type, \eqref{primi_prob} is instantiated by \eqref{plain_T_Prim}, \eqref{T_SQ}, or \eqref{T_SQIW}. $p_{o}\left ( x \right )$ refers to the overall probability of occupancy at point x. Correspondingly, the overall probability of being free at point x is $p_{free}\left ( x \right ) = 1-p_{o}\left ( x \right )$. In reality, neighboring T-primitives may exhibit correlated occupancy responses. Such correlations may cause $p_{o}\left ( x \right )$ to deviate from the exact occupancy probability because the product term in \eqref{geo_pred} neglects dependencies among $p\left(x|P^{T}_{i}\right )$, with higher-order dependencies similarly neglected when multiple primitives are involved. Therefore, \eqref{geo_pred} is used as a tractable factorized approximation rather than an exact independence model, avoiding explicit modeling of dependencies among T-primitives.

\subsubsection{T-Mixture Model Based Semantic Prediction}
A set of T-primitives can be considered as a T-mixture model, and the final semantic predictions can be calculated via a weighted aggregation of semantic probabilities from all contributing T-primitives in the scene as:
\begin{equation}
    p_{sem}\left ( x \right )  =\frac{ {\textstyle \sum_{i=1}^{K}p\left ( x|P^{T}_{i} \right )a_{i}\tilde{c}_{i} } }{ {\textstyle \sum_{j=1}^{K}p\left ( x|P^{T}_{j} \right)a_{j}}} 
    \label{sem_prob}
\end{equation}
where $\tilde{c}_{i}$ represents the softmax-normalized semantic property of $i$-th T-primitive, and $a_{i}$ refers to the $l1$ normalized opacity property of the $i$-th T-primitive; it acts as a normalized mixture weight that determines the relative semantic contribution of each primitive in \eqref{sem_prob}. 
The weighted semantic sum is normalized only when the total accumulated semantic weight exceeds \(\epsilon=10^{-6}\). Otherwise, a uniform fallback is used, \(p_{\mathrm{sem}}(\mathbf{x}) =\frac{1}{C}\mathbf{1}_C =\left[\frac{1}{C},\frac{1}{C},\ldots,\frac{1}{C}\right],\) where \(C\) denotes the number of occupied semantic classes. This $\epsilon$-guard avoids division by zero and ensures numerical stability. The resulting \(p_{\mathrm{sem}}(\mathbf{x})\) represents the semantic distribution conditioned on occupancy. Accordingly, the probability of occupied class \(c\) is
\(p_o(\mathbf{x})p_{\mathrm{sem},c}(\mathbf{x})\), while the free-space probability is \(1-p_o(\mathbf{x})\). When no primitive contributes, \(p_o(\mathbf{x})=0\), so the uniform semantic fallback does not assign any occupied-class probability and the voxel remains entirely free.

\subsubsection{Student-T-Distribution Based Plain Primitive}
The probability function $p\left ( x|P^{T}_{i} \right )$ of the plain T-distribution primitive with the lowest geometric flexibility can be formulated as:
{\small
\begin{align}
    p\left ( x|P^{T}_{i} \right ) 
    = \left [ 1+\frac{1}{\nu_{i}}\left ( x-m_{i} \right )^{T}\Sigma^{-1}\left( x-m_{i} \right )\right]^{-\frac{\nu_{i}+3}{2}}
\label{plain_T_Prim}
\end{align}}
Compared with the corresponding Gaussian kernel, the Student’s-t-shaped occupancy kernel exhibits heavier tails within its finite rasterization support, leading to better robustness against outliers.

\subsubsection{Student-T-Distribution Based SuperQuadric Primitive}
The T-probabilistic modeling mechanism is adopted to convert superquadrics into occupancy probabilities. We first transform the 3D point position $(x,y,z)$, which is occupied by the $i$-th T-Superquadric primitive, into its local coordinate system as:
\begin{equation}
    \begin{bmatrix}
         X_{S_i}\\
         Y_{S_i}\\
         Z_{S_i}\\
        \end{bmatrix} = R_{i}\times \begin{bmatrix}
         x-m_{x_i}\\
         y-m_{y_i}\\
         z-m_{z_i}\\
        \end{bmatrix}
\end{equation}
where $(X_{S_i},Y_{S_i},Z_{S_i})$ refers to the local coordinate under the $i$-th T-Superquadric primitive, and $R_{i}$ is the rotation matrix of the $i$-th T-Superquadric primitive. Then, the occupancy probability of the 3D point $(x,y,z)$ associated with the $i$-th T-Superquadric primitive can be computed as:
{\small
\begin{equation}
    f_{i}\left ( X_{S_i},Y_{S_i},Z_{S_i} \right ) 
    = \left ( \left |\frac{X_{S_i}}{S_{x_i}}\right | ^{\frac{2}{\epsilon_{2_i}}} + \left | \frac{Y_{S_i}}{S_{y_i}}\right|^{\frac{2}{\epsilon_{2_i}}}\right )^{\frac{\epsilon_{2_i}}{\epsilon_{1_i}}} + \left|\frac{Z_{S_i}}{S_{z_i}}\right|^{\frac{2}{\epsilon_{1_i}}}   
\end{equation}
}
{\small
\begin{equation}
    p\left ( x|P^{T}_{SQi} \right ) =\left(1+\frac{1}{\nu_{i} }\cdot f_{i}\left(X_{S_i},Y_{S_i},Z_{S_i}\right)\right )^{-\frac{\nu_{i}+3}{2}}  
\label{T_SQ}
\end{equation}
}where $(S_{x_i},S_{y_i},S_{z_i})$ refers to the scale parameters of $i$-th T-Superquadric primitive along the $X, Y, Z$ axes. Given $S_{x_i},S_{y_i},S_{z_i} > 0$, $\epsilon_{1_i}, \epsilon_{2_i} > 0$, and $\nu_{i} \geq 7$, we have 
\begin{equation}
    f_{i}\left ( X_{S_i},Y_{S_i},Z_{S_i} \right ) \ge 0
\end{equation}
Therefore, 
\begin{equation}
    1 + \frac{f_{i}\left ( X_{S_i},Y_{S_i},Z_{S_i} \right )}{\nu_{i}}  \ge 1  
\end{equation}
and because 
\begin{equation}
    -\frac{\nu_{i}+3}{2} < 0     
\end{equation}
it follows immediately that $0 \leq p\left ( x|P^{T}_{SQi} \right )\le 1$.

\subsubsection{Student-T-Distribution Based SuperQuadric With Inverse-Warp Primitive}
So far, thanks to the T-distribution kernel enhancement, the T-Superquadric primitive is robust to outliers. Nevertheless, due to limited shape variety, single T-Superquadric primitive cannot capture non-convex, asymmetric shapes, which are very common in the daily 3D driving scene. To solve this limitation, we make the T-Superquadric primitive deformable through inverse-warp enhancement. Specifically, we choose 24 basis field functions $\left \{ B_{k}(u,v,w) \right \}_{k=1}^{24}$ to deform the local coordinates $(X_{S_i},Y_{S_i},Z_{S_i})$ as follows:
{\small
\begin{equation}
    u=\frac{X_{S_i}}{S_{x_i}}, v=\frac{Y_{S_i}}{S_{y_i}}, w=\frac{Z_{S_i}}{S_{z_i}}
\end{equation}
\begin{equation}
    \begin{bmatrix}
         \tilde{X}_{S_i}\\
         \tilde{Y}_{S_i}\\
         \tilde{Z}_{S_i}\\
        \end{bmatrix} = \begin{bmatrix}
         X_{S_i}\\
         Y_{S_i}\\
         Z_{S_i}\\
        \end{bmatrix} - \sum_{k=1}^{24}\omega_{i,k} \cdot B_{k}\left ( u,v,w \right ) 
        \label{deform}
\end{equation}
}since the deformation is represented as a residual coordinate warp, the bounded coefficients $\left \{ \omega_{i,k} \right \} _{k=1}^{24}\in \left [ -1,1 \right ]$ constrain the deformation magnitude but do not by themselves guarantee global invertibility. We therefore do not assume that Eq. \eqref{deform} defines a strict diffeomorphism or guarantees folding-free deformation. Instead, the residual warp is used as a local coordinate perturbation within each primitive's finite rasterization support for occupancy evaluation.
Then, the occupancy probability of the 3D point associated with $i$-th T-Superquadric-Inverse-Warp primitive can be calculated as follows:
{\small
\begin{equation}
    f_{i}\left ( \tilde{X}_{S_i},\tilde{Y}_{S_i},\tilde{Z}_{S_i} \right ) = \left ( \left |\frac{\tilde{X}_{S_i}}{S_{x_i}}\right|^{\frac{2}{\epsilon_{2_i}}} + \left | \frac{\tilde{Y}_{S_i}}{S_{y_i}}\right|^{\frac{2}{\epsilon_{2_i}}}\right )^{\frac{\epsilon_{2_i}}{\epsilon_{1_i}}} + \left|\frac{\tilde{Z}_{S_i}}{S_{z_i}}\right|^{\frac{2}{\epsilon_{1_i}}}  
\end{equation}
}
{\small
\begin{equation}
    p\left ( x|P^{T}_{SQIWi} \right ) =\left(1+\frac{1}{\nu_{i} }\cdot f_{i}\left(\tilde{X}_{S_i},\tilde{Y}_{S_i},\tilde{Z}_{S_i}\right)\right )^{-\frac{\nu_{i}+3}{2}}  
\label{T_SQIW}
\end{equation}
}following the same mathematical deduction as in \eqref{T_SQ}, we can also obtain the result that $0 \leq p\left ( x|P^{T}_{SQIWi} \right )\le 1$.

\begin{table}[htbp]
  \centering
  \begin{adjustbox}{width=0.9\columnwidth}
  \begin{tabular}{c|c|c|c}
    \toprule
    Basis Field & $u$ & $v$ & $w$ \\
    \midrule
    $B_{1}$ & 1.0 & 0.0 & 0.0 \\
    $B_{2}$ & 0.0 & 1.0 & 0.0 \\
    $B_{3}$ & 0.0 & 0.0 & 1.0 \\
    $B_{4}$ & $u$ & 0.0 & 0.0 \\
    $B_{5}$ & 0.0 & $v$ & 0.0 \\
    $B_{6}$ & 0.0 & 0.0 & $w$ \\
    $B_{7}$ & $v$ & 0.0 & 0.0 \\
    $B_{8}$ & $w$ & 0.0 & 0.0 \\
    $B_{9}$ & 0.0 & $w$ & 0.0 \\
    $B_{10}$ & 0.0 & $u$ & 0.0 \\
    $B_{11}$ & 0.0 & 0.0 & $u$ \\
    $B_{12}$ & 0.0 & 0.0 & $v$ \\
    $B_{13}$ & $-w*v$ & $w*u$ & 0.0 \\
    $B_{14}$ & 0.0 & $-u*w$ & $u*v$ \\
    $B_{15}$ & $v*w$ & 0.0 & $-v*u$ \\
    $B_{16}$ & $w^{2}$ & 0.0 & 0.0 \\
    $B_{17}$ & 0.0 & $w^{2}$ & 0.0 \\
    $B_{18}$ & 0.0 & 0.0 & $u^{2}+v^{2}$ \\
    $B_{19}$ & $u^{2}$ & 0.0 & 0.0 \\
    $B_{20}$ & 0.0 & $v^{2}$ & 0.0 \\
    $B_{21}$ & 0.0 & 0.0 & $w^{2}$ \\
    $B_{22}$ & $(u^{2} + v^{2})*u$ & $(u^{2} + v^{2})*v$ & 0.0 \\
    $B_{23}$ & $u*v$ & $u*v$ & 0.0 \\
    $B_{24}$ & $u*v^{2}$ & $u^{2}*v$ & 0.0 \\
    \bottomrule
  \end{tabular}
  \end{adjustbox}
  \caption{\small The 24 basis field functions make the T-Superquadric deformable. Those functions enable full shears, twists, bends, quadratics, radial bulge, and smooth corners.}
  \label{basis_field}
\end{table}
As summarized in Table I, our basis fields, motivated by classical global deformation and free-form deformation studies \cite{Barr,Sederberg,Terzopoulos}, are designed to enable more expressive geometric modeling. Specifically, $B_{1}$ to $B_{12}$ form a complete basis for 3D affine displacement fields, producing effects including translation, scaling, shearing, and rotation. $B_{13}$ to $B_{15}$ and $B_{16}$ to $B_{17}$ capture low-order polynomial modes associated with twisting and bending \cite{Barr}, respectively, while $B_{18}$ to $B_{24}$ introduce higher-order polynomial shape variations, including radial bulging and mixed-coordinate asymmetric deformations.

\subsection{Transformer}
\subsubsection{Fused Depth Map Guided 3D Deformable Attention}
\begin{figure}[htbp]
     \centering
        \includegraphics[width=\columnwidth]{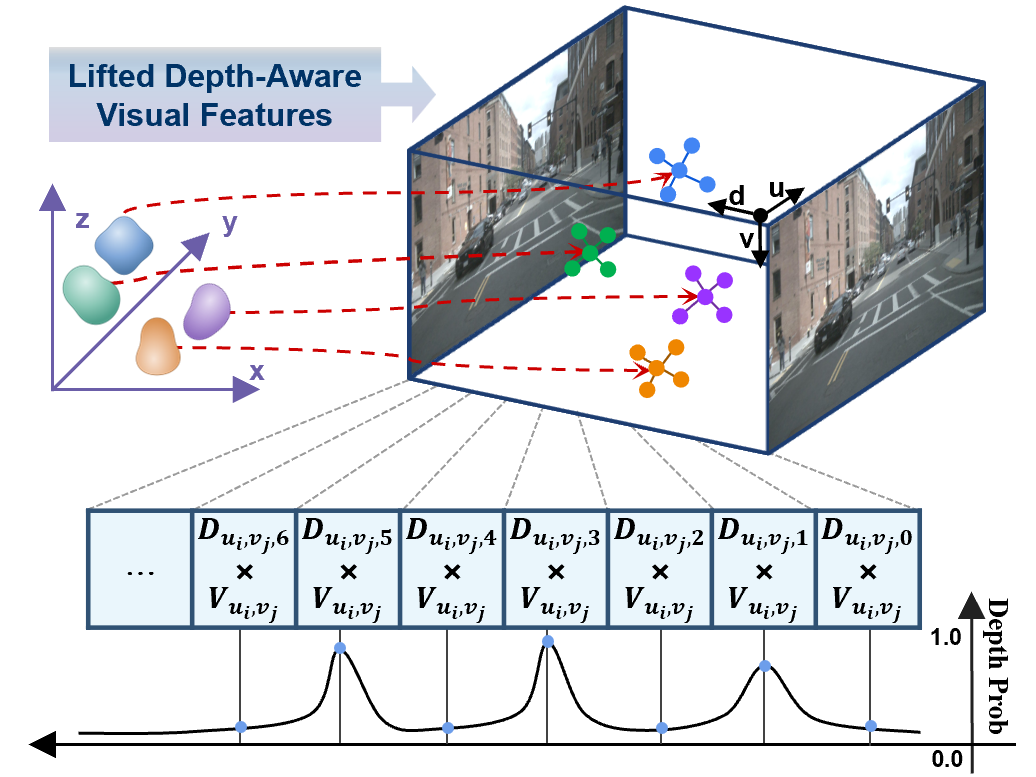}
        \caption{\small \textbf{Fused Depth Map Guided 3D Deformable Attention} For the same visual feature at location $(u_{i},v_{j})$, several different depth values ranging between $[0.0, 1.0]$ along the depth-axis are used to weight the same pixel feature, resulting in the depth encoded visual feature to mitigate the projection ambiguity problem stemming from the projection process.}
        \label{3dfa}
\end{figure}
\begin{figure*}[htpb]
     \centering
        \includegraphics[width=\textwidth]{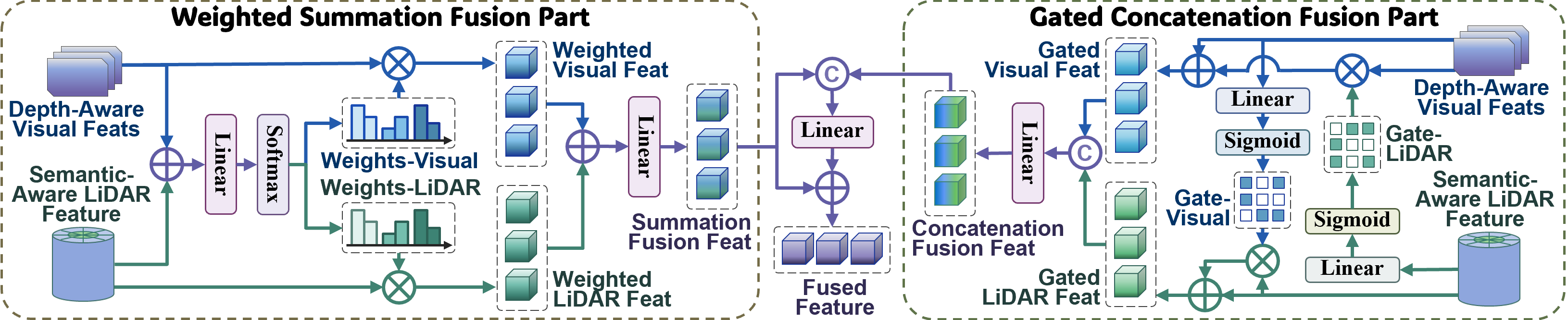}
        \caption{\small \textbf{Gate-Concatenation and Weight-Summation Fusion Module.} The left part exhibits the multi-modality weighted summation fusion part, and the right part exhibits the multi-modality gated concatenation fusion part. The outputs of the two major parts are further fused to produce the final fused feature for each T-Primitive.}
        \label{gcwsfusion}
\end{figure*}
Taking multi-scale lifted depth-aware visual features and a set of T-primitive as input, we use a 3D deformable attention operator, namely 3DFA \cite{dfa3d}, which performs feature aggregation on lifted visual features to mitigate the 3D to 2D projection ambiguity problem, to aggregate visual features for each T-primitive. As shown in Fig.~\ref{TFusionOcc} camera branch and Fig.~\ref{3dfa}, the multi-scale lifted depth-aware visual features $Vol^{Cam}_{depth}$ are obtained through outer product operation $F_{depth}^{Cam} \otimes V_{depth}^{fuse}$, this operation gives an accurate depth encoding for each visual feature in the feature map. 
For each T-primitive feature sampling $p^{T}_{i}$, we first use the mean location $m_{i}$ of each $p^{T}_{i}$ as the center, convert the center location from the cylindrical coordinate to the cartesian coordinate, and feed its associate query feature $q_{i}$ to an MLP layer, resulting in 13 offsets $\Delta m_{j}$ with respect to the center to form a set of 3D sampling points $m^{samp}=\left \{ m^{samp}_{i} = m_{i} + \Delta m_{j}\mid j=1,...,13 \right \} $. Then, we leverage extrinsic and intrinsic matrices to project each 3D sampling point $m^{samp}_{i}$ onto the lifted depth-aware visual features, resulting in the location of the sampling of the pixel frame $\bar{m}^{samp}_{i} = \left ( u_{i},v_{i},d_{i} \right ) $. Lastly, a 3D deformable attention operator 3DFA takes $\bar{m}^{samp}_{i}$ as a sampling location to perform cross-attention visual feature aggregation on $Vol^{Cam}_{depth}$, resulting in an updated query feature $q_{i}^{updated}$. The overall sampling process can be formulated as follows.
\begin{equation}
    q_{i}^{updated} = \sum_{i=1}^{13} 3DFA\left ( q_{i},\bar{m}^{samp}_{i},Vol^{Cam}_{depth} \right ) 
\end{equation}
After the 3D cross-attention process, $q_{i}^{updated}$ already aggregates sufficient depth-aware visual features and is passed to the GCWSFusion module for feature fusion. Meanwhile, the set of 3D sampling points $m^{samp}$ for each T-primitive is also passed on to the GCWSFusion module for semantic-aware LiDAR feature aggregation.

\subsubsection{Gate-Concatenation and Weight-Summation Fusion (GCWSFusion) Module}
Given a set of 3D sampling points $m^{samp}$ for each T-primitive, we apply bilinear interpolation on $Vol^{Lidar}_{sem}$ to aggregate a semantic-aware LiDAR feature for each T-primitive. Then, both the semantic-aware LiDAR feature and the depth-aware visual feature of each T-primitive are fed into the GCWSFusion module for late-stage feature fusion. The detailed implementation of GCWSFusion module is demonstrated in Fig.~\ref{gcwsfusion}. It consists of two central parts, the weighted summation fusion part, as highlighted in the green rectangle on the left of Fig.~\ref{gcwsfusion}, and the gated concatenation fusion part, as highlighted in the pink rectangle on the right of Fig.~\ref{gcwsfusion}. Both parts take the same inputs, but perform different feature fusion strategies. 

For the weighted summation fusion part, the semantic-aware LiDAR feature and the depth-aware visual feature are summed in elements first, then passed through several linear layers followed by a softmax layer, resulting in two weight values, $W_{Lidar}$ and $W_{Vis}$. The $W_{Lidar}$ and $W_{Vis}$ are multiplied again with the corresponding semantic-aware LiDAR feature and depth-aware visual feature, respectively, resulting in the weighted LiDAR feature and the weighted visual feature. Lastly, the weighted LiDAR feature and the weighted visual feature are summed and passed through an extra set of linear layers, yielding the summation fusion feature. 
This design adaptively weights the dominance information across two sensors.

For the gated concatenation fusion part, the semantic-aware LiDAR feat and the depth-aware visual feat are passed through several linear layers, followed by a sigmoid function, respectively, resulting in a visual gate vector $G_{Vis}$ and a LiDAR gate vector $G_{Lidar}$. $G_{Vis}$ is element-wise multiplied with the semantic-aware LiDAR feature, followed by a skip connection that results in the gated LiDAR feature. $G_{Lidar}$ is element-wise multiplied with the depth-aware visual feature, followed by a skip connection that results in the gated visual feature. The gated LiDAR and visual features are concatenated along the feature channel dimension, followed by another set of linear layers that reduces the fused feature channel dimension, yielding the concatenation fusion feature. This design can dynamically amplify or suppress feature information within the two different modalities, thereby achieving denoising and information enhancement during the feature fusion process.

Lastly, the summation and concatenation fusion features are combined via feature-channel concatenation, followed by extra linear layers, yielding the final fused feature.

\subsubsection{3D Sparse Conv-based Self-Attention Module}
Each T-primitive center $m_{i}$ is treated as a 3D point, and we can acquire a set of 3D points based on a set of T-primitives. Then, this T-primitive-based sparse 3D point cloud undergoes a coordinate transformation from Cartesian to Cylindrical coordinates, followed by voxelization, yielding a sparse voxel volume in cylindrical coordinates. Lastly, the 3D sparse convolution applies to this sparse voxel volume to perform the self-attention operation. Since our T-primitives are initialized in cylindrical coordinates, we perform the self-attention operation in the same coordinates to better preserve the scene's fine-grained 3D geometry.

\subsubsection{T-Primitive Property Refinement}
As presented in \eqref{artri_update}, the physical attributes of the three proposed T-Primitives are refined by processing each associated query vector through an MLP to produce updated attributes, which are subsequently added to the original attributes for refinement.
{\small
\begin{flalign}
& (\Delta{m}_i,\hat{s}_i,\hat{r}_i,\hat\alpha_{i},\hat{c}_i) = MLP(Q_{P_i}) & \notag \\
& (\Delta{m}_i,\hat{s}_i,\hat{r}_i,\hat\alpha_{i},\hat{\epsilon}_{1_i},\hat{\epsilon}_{2_i},\hat{c}_i) = MLP(Q_{SQ_i}) & \notag \\
& (\Delta{m}_i,\hat{s}_i,\hat{r}_i,\hat\alpha_{i},\hat{\epsilon}_{1_i},\hat{\epsilon}_{2_i},\left \{ \hat{\omega_{i,k}} \right \}_{k=1}^{24},\hat{c}_i) = MLP(Q_{SQIW_i}) & \notag \\
& \hat{P^{T}_{i}} = (m_{i}+\Delta{m}_i,\hat{s}_i,\hat{r}_i,\hat\alpha_{i},\hat{c}_i) & \notag \\
& \hat{P^{T}_{SQ_i}} = (m_{i}+\Delta{m}_i,\hat{s}_i,\hat{r}_i,\hat\alpha_{i},
\hat{\epsilon}_{1_i},\hat{\epsilon}_{2_i},\hat{c}_i) & \notag \\
& \hat{P^{T}_{SQIW_i}} = (m_{i}+\Delta{m}_i,\hat{s}_i,\hat{r}_i,\hat\alpha_{i},
\hat{\epsilon}_{1_i},\hat{\epsilon}_{2_i},\left \{ \hat{\omega_{i,k}} \right \}_{k=1}^{24},\hat{c}_i) &
\label{artri_update}
\end{flalign}}

\section{Experimental Results}\label{simulation}
\subsection{Implementation Details}
The TFusionOcc model takes six surround-view images and 10 LiDAR sweeps per data sample and leverages ResNet50-DCN \cite{residual} as a 2D Backbone, initialized with weights from FCOS3D \cite{fcos3d}. The transformer layer is stacked iteratively four times. For T-primitive initialization, we use 19200 LiDAR anchors and 6400 camera anchors, resulting in 25600 primitive queries in the full setting. The scale parameters $s_{i}$ are constrained to $[0.01, 1.8]$, the Superquadric exponents $\epsilon_{1_i},\epsilon_{2_i}$ are constrained to $[0.1, 2.0]$, and the inverse-warp coefficients satisfy $\omega_{i,k}\in \left [ -1.0,1.0 \right ] $. During splatting, each primitive is evaluated only within its clipped axis-aligned rasterization window and contributes zero outside this support. The T-P, T-SQ, and T-SQ-IW kernels are evaluated according to Eqs. \eqref{plain_T_Prim}, \eqref{T_SQ}, and \eqref{T_SQIW}, respectively, followed by the occupancy and semantic aggregation in Eqs. \eqref{geo_pred} and \eqref{sem_prob}. AdamW optimizer is used, with an initial learning rate of 2e-4 and a weight decay of 0.01. The learning rate is decayed using a multistep scheduler. The predicted occupancy has a resolution of $200 \times 200 \times 16$ for full-scale evaluation on SurroundOcc-nuScenes, Occ3D-nuScenes, and nuScenes-C datasets and $256 \times 256 \times 32$ for full-scale evaluation on semanticKITTI. For data augmentation, we employ photometric distortions and a grid mask on the input images during training. The model is implemented using MMDetection3D \cite{mmdet3d} and trained for 24 epochs with a batch size of 4 on 2 NVIDIA H100 GPUs.

\subsection{Loss Function}
To train TFusionOcc, we take advantage of the Lovász-Softmax loss \cite{lovasz} $L_{lovasz}$ and the binary cross-entropy loss $L_{BCE}$ to optimize the 3D semantic occupancy task. For depth supervision, a binary cross-entropy loss $L^{depth}_{BCE}$ is used to refine the depth distribution feature. The final loss is formulated as: 
\begin{equation}
    L_{Total} = L_{lovasz} + \lambda \cdot L_{BCE} + L_{BCE}^{depth}
\end{equation}
where $\lambda$ balance the loss weight, in our case, we set $\lambda=10$. The final loss is applied only to the output of the last iterated transformer.

\subsection{Dataset}
The nuScenes dataset \cite{nuscenes} serves as the primary dataset for our experiments. We additionally use nuScenes-C \cite{nuscenes-c} to evaluate robustness to camera corruptions, including snow, fog, brightness variations, frame loss, and camera failures. To further assess the applicability of TFusionOcc beyond the nuScenes domain, we also evaluate it on the SemanticKITTI dataset \cite{semantickitti}.

For 3D semantic occupancy prediction on nuScenes, we use the dense annotations provided by SurroundOcc \cite{surroundocc} and Occ3D \cite{occ3d}, and evaluate on their validation sets. For SurroundOcc-nuScenes, the perception range is [-50, 50] m along the X and Y axes and [-5, 3] m along the Z axis in LiDAR coordinates. For Occ3D-nuScenes, the corresponding ranges are [-40, 40] m, [-40, 40] m, and [-1, 5.4] m in ego coordinates. Both settings use $1600\times900$ input images and an occupancy resolution of $200\times200\times16$. SurroundOcc and Occ3D contain 17 and 18 semantic classes, respectively, with label 0 denoting free space. For semantic scene completion on SemanticKITTI, we use a perception range of [0, 51.2] m, [-25.6, 25.6] m, and [-2, 4.4] m along the X, Y, and Z axes, respectively, in LiDAR coordinates. The input image resolution is $1241\times376$, and the output occupancy resolution is $256\times256\times32$ over 19 semantic classes.

\subsection{Model Performance Analysis}
To demonstrate the performance of our proposed TFusionOcc, we evaluated it on the SurroundOcc-nuScenes and Occ3D-nuScenes benchmarks. For each benchmark, we evaluate our model across three primitive settings, including T-P (plain T-primitive), T-SQ (T-superquadric), and T-SQ-IW (deformable T-superquadric or T-superquadric with inverse-warp), with 12800 and 25600 total primitives, respectively. The results of the experiment on the SurroundOcc-nuScenes and Occ3D-nuScenes benchmarks are shown in Table \ref{surroundocc-nuscenes} and Table \ref{occ3d-nuscenes}, respectively.
\begin{table*}[t]
  \centering
  \begin{adjustbox}{width=\textwidth}
  \begin{tabular}{c|c|c|c|c|cc|ccccccccccccccccc}
    \toprule
    Method & {\begin{tabular}[c]{@{}c@{}} Multi\\ Sup\end{tabular}} & {\begin{tabular}[c]{@{}c@{}} Lidar\\ Sweep\end{tabular}} & 2D-Backbone & Modality & IoU & mIoU & \rotatebox{90}{\textcolor{barrier}{$\bullet$} barrier} & \rotatebox{90}{\textcolor{bicycle}{$\bullet$} bicycle} & \rotatebox{90}{\textcolor{bus}{$\bullet$} bus} & \rotatebox{90}{\textcolor{car}{$\bullet$} car} & \rotatebox{90}{\textcolor{construction}{$\bullet$} const. veh.} & \rotatebox{90}{\textcolor{motorcycle}{$\bullet$} motorcycle} & \rotatebox{90}{\textcolor{pedestrian}{$\bullet$} pedestrian} & \rotatebox{90}{\textcolor{cone}{$\bullet$} traffic cone} & \rotatebox{90}{\textcolor{trailer}{$\bullet$} trailer} & \rotatebox{90}{\textcolor{truck}{$\bullet$} truck} & \rotatebox{90}{\textcolor{driveable}{$\bullet$} drive. surf.} & \rotatebox{90}{\textcolor{flat}{$\bullet$} other flat} & \rotatebox{90}{\textcolor{sidewalk}{$\bullet$} sidewalk} & \rotatebox{90}{\textcolor{terrain}{$\bullet$} terrain} & \rotatebox{90}{\textcolor{manmade}{$\bullet$} manmade} & \rotatebox{90}{\textcolor{vegetation}{$\bullet$} vegetation} \\
     \midrule 
    MonoScene \cite{monoscene} & \XSolidBrush & - & R101-DCN & C & 23.96 & 7.31 & 4.03 & 0.35 & 8.00 & 8.04 & 2.90 & 0.28 & 1.16 & 0.67 & 4.01 & 4.35 & 27.72 & 5.20 & 15.13 & 11.29 & 9.03 & 14.86\\
    Atlas \cite{atlas} & \XSolidBrush & - & C & 28.66 & 15.00 & 10.64 & 5.68 & 19.66 & 24.94 & 8.90 & 8.84 & 6.47 & 3.28 & 10.42 & 16.21 & 34.86 & 15.46 & 21.89 & 20.95 & 11.21 & 20.54 \\
    BEVFormer \cite{bevformer} & \XSolidBrush & - & R101-DCN & C & 30.50 & 16.75 & 14.22 & 6.58 & 23.46 & 28.28 & 8.66 & 10.77 & 6.64 & 4.05 & 11.20 & 17.78 & 37.28 & 18.00 & 22.88 & 22.17 & 13.80 & 22.21 \\
    TPVFormer \cite{tpvformer} & \XSolidBrush & - & R101-DCN & C & 30.86 & 17.10 & 15.96 & 5.31 & 23.86 & 27.32 & 9.79 & 8.74 & 7.09 & 5.20 & 10.97 & 19.22 & 38.87 & 21.25 & 24.26 & 23.15 & 11.73 & 20.81 \\
    C-CONet \cite{openoccupancy} & \XSolidBrush & - & R101 & C & 26.10 & 18.40 & 18.60 & 10.00 & 26.40 & 27.40 & 8.60 & 15.70 & 13.30 & 9.70 & 10.90 & 20.20 & 33.00 & 20.70 & 21.40 & 21.80 & 14.70 & 21.30 \\
    InverseMatrixVT3D \cite{inversematrixvt3d} & \XSolidBrush & - & R101-DCN & C & 30.03 & 18.88 & 18.39 & 12.46 & 26.30 & 29.11 & 11.00 & 15.74 & 14.78 & 11.38 & 13.31 & 21.61 & 36.30 & 19.97 & 21.26 & 20.43 & 11.49 & 18.47 \\
    OccFormer \cite{occformer} & \XSolidBrush & - & R101-DCN & C & 31.39 & 19.03 & 18.65 & 10.41 & 23.92 & 30.29 & 10.31 & 14.19 & 13.59 & 10.13 & 12.49 & 20.77 & 38.78 & 19.79 & 24.19 & 22.21 & 13.48 & 21.35 \\
    FB-Occ \cite{fbocc} & \XSolidBrush & - & R101 & C & 31.50 & 19.60 & 20.60 & 11.30 & 26.90 & 29.80 & 10.40 & 13.60 & 13.70 & 11.40 & 11.50 & 20.60 & 38.20 & 21.50 & 24.60 & 22.70 & 14.80 & 21.60 \\
    RenderOcc \cite{renderocc} & \XSolidBrush & - & R101 & C & 29.20 & 19.00 & 19.70 & 11.20 & 28.10 & 28.20 & 9.80 & 14.70 & 11.80 & 11.90 & 13.10 & 20.10 & 33.20 & 21.30 & 22.60 & 22.30 & 15.30 & 20.90 \\
    GaussianFormer \cite{gaussianformer} & \XSolidBrush & - & R101-DCN & C & 29.83 & 19.10 & 19.52 & 11.26 & 26.11 & 29.78 & 10.47 & 13.83 & 12.58 & 8.67 & 12.74 & 21.57 & 39.63 & 23.28 & 24.46 & 22.99 & 9.59 & 19.12 \\
    Co-Occ \cite{co-occ} & \XSolidBrush & - & R101 & C & 30.00 & 20.30 & 22.50 & 11.20 & 28.60 & 29.50 & 9.90 & 15.80 & 13.50 & 8.70 & 13.60 & 22.20 & 34.90 & 23.10 & 24.20 & 24.10 & 18.00 & 24.80 \\
    GaussianFormer2-256 \cite{gaussianformer2} & \XSolidBrush & - & R101-DCN & C & 31.74 & 20.82 & 21.39 & 13.44 & 28.49 & 30.82 & 10.92 & 15.84 & 13.55 & 10.53 & 14.04 & 22.92 & 40.61 & 24.36 & 26.08 & 24.27 & 13.83 & 21.98 \\
    QuadricFormer \cite{quadricformer} & \XSolidBrush & - & R101-DCN & C & 31.22 & 20.12 & 19.58 & 13.11 & 27.27 & 29.64 & 11.25 & 16.26 & 12.65 & 9.15 & 12.51 & 21.24 & 40.20 & 24.34 & 25.69 & 24.24 & 12.95 & 21.86 \\
    SurroundOcc \cite{surroundocc} & \XSolidBrush & - & R101-DCN & C & 31.49 & 20.30 & 20.59 & 11.68 & 28.06 & 30.86 & 10.70 & 15.14 & 14.09 & 12.06 & 14.38 & 22.26 & 37.29 & 23.70 & 24.49 & 22.77 & 14.89 & 21.86 \\
    Inverse++ \cite{inverse++} & \XSolidBrush & - & R101-DCN & C & 31.73 & 20.91 & 20.90 & 13.27 & 28.40 & 31.37 & 11.90 & 17.76 & 15.39 & 13.49 & 13.32 & 23.19 & 39.37 & 22.85 & 25.27 & 23.68 & 13.43 & 20.98 \\
    \hline
    LMSCNet \cite{lmscnet} & \XSolidBrush & - & - & L & 36.60 & 14.90 & 13.10 & 4.50 & 14.70 & 22.10 & 12.60 & 4.20 & 7.20 & 7.10 & 12.20 & 11.50 & 26.30 & 14.30 & 21.10 & 15.20 & 18.50 & 34.20 \\
    L-CONet \cite{openoccupancy} & \XSolidBrush & 10 & - & L & 39.40 & 17.70 & 19.20 & 4.00 & 15.10 & 26.90 & 6.20 & 3.80 & 6.80 & 6.00 & 14.10 & 13.10 & 39.70 & 19.10 & 24.00 & 23.90 & 25.10 & 35.70 \\
    Co-Occ \cite{co-occ} & \XSolidBrush & 10 & - & L & 42.20 & 22.90 & 22.00 & 6.90 & 25.70 & 32.40 & 14.50 & 13.50 & 21.00 & 10.50 & 18.00 & 22.50 & 36.60 & 21.80 & 24.60 & 25.70 & 31.20 & 39.90 \\
    Co-Occ \cite{co-occ} & \XSolidBrush & 10 & R101-DCN & C+L & 41.10 & 27.10 & 28.10 & 16.10 & 34.00 & 37.20 & 17.00 & 21.60 & 20.80 & 15.90 & 21.90 & 28.70 & 42.30 & 25.40 & 29.10 & 28.60 & 28.20 & 38.00 \\
    GaussianFormer3D \cite{gaussianformer3d} & \XSolidBrush & 10 & R101-DCN & C+L & 43.30 & 27.10 & 26.90 & 15.80 & 32.70 & 36.10 & 18.60 & 21.70 & 24.10 & 13.00 & 21.30 & 29.00 & 40.60 & 23.70 & 27.30 & 28.20 & 32.60 & 42.30 \\
    OccFusion \cite{occfusion} & \XSolidBrush & 10 & R101-DCN & C+R & 33.97 & 20.73 & 20.46 & 13.98 & 27.99 & 31.52 & 13.68 & 18.45 & 15.79 & 13.05 & 13.94 & 23.84 & 37.85 & 19.60 & 22.41 & 21.20 & 16.16 & 21.81 \\
    OccFusion \cite{occfusion} & \XSolidBrush & 10 & R101-DCN & C+L & 43.53 & 27.55 & 25.15 & 19.87 & 34.75 & 36.21 & 20.03 & 23.11 & 25.25 & 17.50 & 22.70 & 30.06 & 39.47 & 23.26 & 25.68 & 27.57 & 29.54 & 40.60 \\
    OccFusion \cite{occfusion} & \XSolidBrush & 10 & R101-DCN & C+L+R & 43.96 & 28.27 & 28.32 & 20.95 & 35.06 & 36.84 & 20.33 & 26.22 & 25.86 & 19.17 & 21.27 & 30.64 & 40.08 & 23.75 & 25.56 & 27.63 & 29.82 & 40.82 \\
    GaussianFusionOcc \cite{gaussianfusionocc} & \XSolidBrush & 10 & R101-DCN & C+L & 45.16 & 30.21 & 30.22 & 18.70 & 35.91 & 39.57 & 22.67 & 27.36 & 30.10 & 18.59 & 24.45 & 31.25 & 43.06 & 25.76 & 29.12 & 29.33 & 34.65 & 42.70 \\
    GaussianFusionOcc \cite{gaussianfusionocc} & \XSolidBrush & 10 & R101-DCN & C+L+R & 45.20 & 30.37 & 30.43 & 18.54 & 36.23 & 39.66 & 22.57 & 27.35 & \textbf{30.30} & 19.14 & \textbf{24.56} & 31.95 & 42.60 & 25.82 & 29.48 & 29.70 & 34.78 & 42.95 \\
    OccCylindrical \cite{occcylindrical} & \XSolidBrush & 10 & Res-50 & C+L & 44.94 & 28.67 & 26.17 & \textbf{22.12} & 31.47 & 36.84 & 17.95 & \textbf{27.77} & 29.85 & \textbf{23.90} & 20.64 & 28.27 & 43.00 & 23.14 & 27.99 & 27.81 & 30.81 & 40.95 \\
    DAOcc \cite{daocc} & \Checkmark & 10 & Res-50 & C+L & 45.00 & 30.50 & 30.80 & 19.50 & 34.00 & 38.80 & \textbf{25.00} & 27.70 & 29.90 & 22.50 & 23.20 & 31.60 & 41.00 & \textbf{25.90} & 29.40 & 29.90 & 35.20 & 43.50 \\
    SDGOcc \cite{sdgocc} & \Checkmark & 10 & Res-50 & C+L & - & \textbf{31.70} & \textbf{31.20} & 15.60 & \textbf{39.70} & \textbf{42.40} & 22.20 & 26.90 & 29.70 & 22.70 & 24.20 & \textbf{33.20} & \textbf{45.90} & 25.10 & \textbf{31.80} & \textbf{33.50} & \textbf{38.90} & \textbf{44.50} \\
    \hline
    \rowcolor{green!30} TFusionOcc (T-SQ-12800) & \XSolidBrush & 10 & Res-50 & C+L & 46.70 & 30.76 & 30.67 & \textcolor{red}{\textbf{21.27}} & 33.12 & 38.96 & 22.09 & 27.54 & 30.47 & 22.87 & 21.65 & 29.86 & 42.63 & \textcolor{blue}{\textbf{27.18}} & 29.80 & 30.91 & 36.81 & 46.26 \\
    \rowcolor{green!30} TFusionOcc (T-SQ-IW-12800) & \XSolidBrush & 10 & Res-50 & C+L & \textcolor{blue}{\textbf{46.92}} & \textcolor{blue}{\textbf{31.21}} & \textcolor{blue}{\textbf{31.65}} & 20.13 & \textcolor{red}{\textbf{34.75}} & \textcolor{blue}{\textbf{39.70}} & \textcolor{blue}{\textbf{23.08}} & 27.00 & 30.89 & 23.33 & 22.02 & \textcolor{blue}{\textbf{30.40}} & \textcolor{blue}{\textbf{43.11}} & \textcolor{red}{\textbf{27.78}} & \textcolor{red}{\textbf{30.25}} & \textcolor{red}{\textbf{31.47}} & \textcolor{blue}{\textbf{37.15}} & \textcolor{blue}{\textbf{46.60}} \\
    \hline
    \rowcolor{red!30} TFusionOcc (T-P-25600) & \XSolidBrush & 10 & Res-50 & C+L & 45.49 & 30.04 & 30.15 & 19.58 & 34.20 & 38.64 & 22.10 & 26.60 & 29.85 & 22.24 & 22.44 & 29.65 & 41.08 & 25.19 & 28.43 & 29.71 & 35.77 & 45.01 \\
    \rowcolor{red!30} TFusionOcc (T-SQ-25600) & \XSolidBrush & 10 & Res-50 & C+L & 46.14 & 30.88 & 30.87 & \textcolor{red}{\textbf{21.26}} & 33.68 & 39.25 & 22.54 & \textcolor{red}{\textbf{28.34}} & \textcolor{red}{\textbf{31.45}} & \textcolor{blue}{\textbf{23.61}} & \textcolor{blue}{\textbf{22.80}} & 30.16 & 41.76 & 26.12 & 29.23 & 30.26 & 36.77 & 46.06 \\
    \rowcolor{red!30} TFusionOcc (T-SQ-IW-25600) & \XSolidBrush & 10 & Res-50 & C+L & \textcolor{red}{\textbf{47.01}} & \textcolor{red}{\textbf{31.47}} & \textcolor{red}{\textbf{31.96}} & 20.78 & \textcolor{blue}{\textbf{34.57}} & \textcolor{red}{\textbf{39.73}} & \textcolor{red}{\textbf{24.02}} & \textcolor{blue}{\textbf{27.82}} & \textcolor{blue}{\textbf{31.19}} & \textcolor{red}{\textbf{23.82}} & \textcolor{red}{\textbf{23.05}} & \textcolor{red}{\textbf{30.67}} & \textcolor{red}{\textbf{43.17}} & \textcolor{blue}{\textbf{27.18}} & \textcolor{blue}{\textbf{30.13}} & \textcolor{blue}{\textbf{31.18}} & \textcolor{red}{\textbf{37.52}} & \textcolor{red}{\textbf{46.71}} \\
    \bottomrule
  \end{tabular}
  \end{adjustbox}
  \caption{\textbf{3D semantic occupancy prediction results on SurroundOcc-nuScenes validation set}. Notation of modality: Camera (C), LiDAR (L), Radar (R).  The best and second-best performance of our algorithm across different settings is indicated in \textcolor{red}{\textbf{bold red}} and \textcolor{blue}{\textbf{bold blue}}, respectively.}
  \label{surroundocc-nuscenes}
\end{table*}
\begin{table*}[t]
  \centering
  \begin{adjustbox}{width=\textwidth}
  \begin{tabular}{c|c|c|c|c|c|c|c|ccccccccccccccccc}
    \toprule
    Method & Mask & {\begin{tabular}[c]{@{}c@{}} Multi\\ Sup\end{tabular}} & {\begin{tabular}[c]{@{}c@{}} Lidar\\ Sweep\end{tabular}} &{\begin{tabular}[c]{@{}c@{}}Hist\\ Frame\end{tabular}} & Backbone & Modality & mIoU & 
    \rotatebox{90}{\textcolor{others}{$\bullet$} others} &
    \rotatebox{90}{\textcolor{barrier}{$\bullet$} barrier} & \rotatebox{90}{\textcolor{bicycle}{$\bullet$} bicycle} & \rotatebox{90}{\textcolor{bus}{$\bullet$} bus} & \rotatebox{90}{\textcolor{car}{$\bullet$} car} & \rotatebox{90}{\textcolor{construction}{$\bullet$} const. veh.} & \rotatebox{90}{\textcolor{motorcycle}{$\bullet$} motorcycle} & \rotatebox{90}{\textcolor{pedestrian}{$\bullet$} pedestrian} & \rotatebox{90}{\textcolor{cone}{$\bullet$} traffic cone} & \rotatebox{90}{\textcolor{trailer}{$\bullet$} trailer} & \rotatebox{90}{\textcolor{truck}{$\bullet$} truck} & \rotatebox{90}{\textcolor{driveable}{$\bullet$} drive. surf.} & \rotatebox{90}{\textcolor{flat}{$\bullet$} other flat} & \rotatebox{90}{\textcolor{sidewalk}{$\bullet$} sidewalk} & \rotatebox{90}{\textcolor{terrain}{$\bullet$} terrain} & \rotatebox{90}{\textcolor{manmade}{$\bullet$} manmade} & \rotatebox{90}{\textcolor{vegetation}{$\bullet$} vegetation} \\
     \midrule 
    MonoScene \cite{monoscene} & \XSolidBrush & \XSolidBrush & - & - & Res-101 & C & 6.06 & 1.75 & 7.23 & 4.26 & 4.93 & 9.38 & 5.67 & 3.98 & 3.01 & 5.90 & 4.45 & 7.17 & 14.91 & 6.32 & 7.92 & 7.43 & 1.01 & 7.65 \\
    BEVDet \cite{bevdet} & \XSolidBrush & \XSolidBrush & - & - & Res-101 & C & 11.73 & 2.09 & 15.29 & 0.0 & 4.18 & 12.97 & 1.35 & 0.0 & 0.43 & 0.13 & 6.59 & 6.66 & 52.72 & 19.04 & 26.45 & 21.78 & 14.51 & 15.26 \\
    BEVFormer \cite{bevformer} & \XSolidBrush & \XSolidBrush & - & - & Res-101 & C & 23.67 & 5.03 & 38.79 & 9.98 & 34.41 & 41.09 & 13.24 & 16.50 & 18.15 & 17.83 & 18.66 & 27.70 & 48.95 & 27.73 & 29.08 & 25.38 & 15.41 & 14.46 \\
    BEVStereo \cite{bevstereo} & \XSolidBrush & \XSolidBrush & - & - & Res-101 & C & 24.51 & 5.73 & 38.41 & 7.88 & 38.70 & 41.20 & 17.56 & 17.33 & 14.69 & 10.31 & 16.84 & 29.62 & 54.08 & 28.92 & 32.68 & 26.54 & 18.74 & 17.49 \\
    TPVFormer \cite{tpvformer} & \XSolidBrush & \XSolidBrush & - & - & Res-101 & C & 28.34 & 6.67 & 39.20 & 14.24 & 41.54 & 46.98 & 19.21 & 22.64 & 17.87 & 14.54 & 30.20 & 35.51 & 56.18 & 33.65 & 35.69 & 31.61 & 19.97 & 16.12 \\
    OccFormer \cite{occformer} & \XSolidBrush & \XSolidBrush & - & - & Res-101 & C & 21.93 & 5.94 & 30.29 & 12.32 & 34.40 & 39.17 & 14.44 & 16.45 & 17.22 & 9.27 & 13.90 & 26.36 & 50.99 & 30.96 & 34.66 & 22.73 & 6.76 & 6.97 \\
    RenderOcc \cite{renderocc} & \XSolidBrush & \XSolidBrush & - & - & Swin-B & C & 26.11 & 4.84 & 31.72 & 10.72 & 27.67 & 26.45 & 13.87 & 18.20 & 17.67 & 17.84 & 21.19 & 23.25 & 63.20 & 36.42 & 46.21 & 44.26 & 19.58 & 20.72 \\
    CTF-Occ \cite{occ3d} & \XSolidBrush & \XSolidBrush & - & - & Res-101 & C & 28.53 & 8.09 & 39.33 & 20.56 & 38.29 & 42.24 & 16.93 & 24.52 & 22.72 & 21.05 & 22.98 & 31.11 & 53.33 & 33.84 & 37.98 & 33.23 & 20.79 & 18.00 \\
    Inverse++ \cite{inverse++} & \XSolidBrush & \XSolidBrush & - & - & Res-101 & C & 31.04 & 9.56 & 41.91 & 23.53 & 42.38 & 46.35 & 18.61 & 28.03 & 26.61 & 24.77 & 25.93 & 34.81 & 60.10 & 33.23 & 37.62 & 34.83 & 19.20 & 20.26 \\
    \hline
    RadOcc \cite{radocc} & \Checkmark & \XSolidBrush & 10 & - & Swin-B & C+L & 49.38 & 10.90 & 58.20 & 25.00 & 57.90 & 62.90 & 34.00 & 33.50 & 50.10 & 32.10 & 48.90 & 52.10 & 82.90 & 42.70 & 55.30 & 58.30 & 68.60 & 66.00 \\
    OccFusion \cite{occfusion} & \Checkmark & \XSolidBrush & 10 & - & R101-DCN & C+L+R & 46.67 & 12.40 & 50.30 & 31.50 & 57.60 & 58.80 & 34.00 & 41.00 & 47.20 & 29.70 & 42.00 & 48.00 & 78.40 & 35.70 & 47.30 & 52.70 & 63.50 & 63.30 \\
    OccFusion \cite{occfusion} & \Checkmark & \XSolidBrush & 10 & - & R101-DCN & C+L & 48.74 & 12.40 & 51.80 & 33.00 & 54.60 & 57.70 & 34.00 & 43.00 & 48.40 & 35.50 & 41.20 & 48.60 & 83.00 & 44.70 & 57.10 & 60.00 & 62.50 & 61.30 \\
    GaussianFormer3D \cite{gaussianformer3d} & \Checkmark & \XSolidBrush & 10 & - & R101-DCN & C+L & 46.40 & 9.80 & 50.00 & 31.30 & 54.00 & 59.40 & 28.10 & 36.20 & 46.20 & 26.70 & 40.20 & 49.70 & 79.10 & 37.30 & 49.00 & 55.00 & 69.10 & 67.60 \\
    RM$^2$Occ \cite{rm2occ} & \Checkmark & \XSolidBrush & 10 & - & R101-DCN & C+L & 47.82 & 13.34 & 54.53 & 33.81 & 58.30 & 59.97 & 34.45 & 34.89 & 29.43 & 31.53 & 39.91 & 42.44 & \textbf{96.01} & 49.76 & 61.28 & \textbf{67.62} & 53.71 & 51.96 \\
    SDGOcc \cite{sdgocc} & \Checkmark & \Checkmark & 10 & - & Res-50 & C+L & 51.66 & 13.20 & 57.80 & 24.30 & 60.30 & 64.30 & 36.20 & 39.40 & 52.40 & 35.80 & \textbf{50.90} & 53.70 & 84.60 & 47.50 & 58.00 & 61.60 & 70.70 & 67.70 \\
    EFFOcc \cite{effocc} & \Checkmark & \XSolidBrush & 10 & - & Res-50 & C+L & 52.82 & 12.09 & 59.67 & 33.39 & 61.76 & 64.98 & 35.46 & 46.01 & 57.09 & 41.04 & 47.87 & 54.59 & 82.76 & 43.95 & 56.37 & 60.23 & \textbf{71.12} & \textbf{69.60} \\
    DAOcc \cite{daocc} & \Checkmark & \Checkmark & 10 & - & Res-50 & C+L & 54.33 & 13.00 & \textbf{60.70} & 39.80 & \textbf{64.00} & \textbf{66.50} & 36.30 & \textbf{49.00} & \textbf{60.10} & 44.30 & 50.70 & \textbf{55.90} & 82.90 & 44.60 & 56.80 & 60.60 & 70.10 & 68.30 \\
    ALOcc \cite{alocc} & \Checkmark & \XSolidBrush & - & 16f & Res-50 & C+D & \textbf{54.46} & \textbf{16.99} & 59.03 & \textbf{40.88} & 58.32 & 64.41 & \textbf{37.22} & 45.90 & 52.68 & \textbf{46.78} & 50.49 & 54.54 & 86.25 & \textbf{51.54} & \textbf{61.69} & 64.81 & 69.12 & 65.12  \\
    \hline
    \rowcolor{green!30} TFusionOcc (T-SQ-12800) & \Checkmark & \XSolidBrush & 10 & - & Res-50 & C+L & 52.34 & 14.82 & 58.85 & 39.48 & 55.10 & 62.24 & \textcolor{blue}{\textbf{35.42}} & 44.19 & \textcolor{blue}{\textbf{58.61}} & 47.17 & \textcolor{blue}{\textbf{44.92}} & 48.43 & 82.54 & 42.87 & 54.65 & 59.52 & \textcolor{blue}{\textbf{71.05}} & \textcolor{red}{\textbf{69.96}} \\
    \rowcolor{green!30} TFusionOcc (T-SQ-IW-12800) & \Checkmark & \XSolidBrush & 10 & - & Res-50 & C+L & \textcolor{blue}{\textbf{52.97}} & \textcolor{red}{\textbf{14.90}} & \textcolor{red}{\textbf{60.59}} & \textcolor{red}{\textbf{40.93}} & \textcolor{blue}{\textbf{56.48}} & \textcolor{red}{\textbf{63.68}} & 35.25 & \textcolor{blue}{\textbf{45.96}} & 58.37 & \textcolor{blue}{\textbf{48.62}} & 44.54 & \textcolor{blue}{\textbf{49.84}} & \textcolor{red}{\textbf{83.42}} & \textcolor{blue}{\textbf{43.85}} & \textcolor{blue}{\textbf{55.09}} & \textcolor{blue}{\textbf{59.67}} & 69.86 & 69.51 \\
    \hline
    \rowcolor{red!30} TFusionOcc (T-P-25600) & \Checkmark & \XSolidBrush & 10 & - & Res-50 & C+L & 51.51 & 14.75 & 58.10 & 37.73 & 51.21 & 62.03 & \textcolor{red}{\textbf{35.55}} & 44.82 & 55.18 & 43.70 & 44.33 & 49.24 & 82.71 & 42.59 & 54.83 & 59.27 & 70.21 & 69.41 \\
    \rowcolor{red!30} TFusionOcc (T-SQ-25600) & \Checkmark & \XSolidBrush & 10 & - & Res-50 & C+L & 51.70 & \textcolor{blue}{\textbf{14.89}} & 59.19 & 37.64 & 53.96 & 62.40 & 33.88 & 44.74 & 58.41 & 47.17 & 41.08 & 47.23 & 82.46 & 43.25 & 54.80 & 59.04 & 70.34 & 68.39 \\
    \rowcolor{red!30} TFusionOcc (T-SQ-IW-25600) & \Checkmark & \XSolidBrush & 10 & - & Res-50 & C+L & \textcolor{red}{\textbf{53.35}} & 14.66 & \textcolor{blue}{\textbf{60.25}} & \textcolor{blue}{\textbf{40.80}} & \textcolor{red}{\textbf{58.39}} & \textcolor{blue}{\textbf{63.67}} & 34.81 & \textcolor{red}{\textbf{45.97}} & \textcolor{red}{\textbf{59.04}} & \textcolor{red}{\textbf{48.72}} & \textcolor{red}{\textbf{45.32}} & \textcolor{red}{\textbf{50.43}} & \textcolor{blue}{\textbf{83.21}} & \textcolor{red}{\textbf{44.70}} & \textcolor{red}{\textbf{55.55}} & \textcolor{red}{\textbf{60.22}} & \textcolor{red}{\textbf{71.41}} & \textcolor{blue}{\textbf{69.75}} \\
    \bottomrule
  \end{tabular}
  \end{adjustbox}
  \caption{\textbf{3D semantic occupancy prediction results on Occ3D-nuScenes benchmark}. Notation of modality: Camera (C), LiDAR (L), Depth Map projected from LiDAR (D), Radar (R). The best and second-best performance of our algorithm across different settings is indicated in \textcolor{red}{\textbf{bold red}} and \textcolor{blue}{\textbf{bold blue}}, respectively.}
  \label{occ3d-nuscenes}
\end{table*}
\begin{table*}[htpb]
\centering
\resizebox{\textwidth}{!}{
\begin{tabular}{c|c|c|c c c c c c|c c c|c c c|c c c c c c c}
\hline
\noalign{\smallskip}
    Methods & Modality & mIoU  
    & \rotatebox{90}{\textcolor{sperson}{$\blacksquare$}  person}
    & \rotatebox{90}{\textcolor{sbicyclist}{$\blacksquare$}  bicyclist}
    & \rotatebox{90}{\textcolor{smotorcyclist}{$\blacksquare$}  motorcyclist.}
    & \rotatebox{90}{\textcolor{sbicycle}{$\blacksquare$}  bicycle}
    & \rotatebox{90}{\textcolor{smotorcycle}{$\blacksquare$} motorcycle}
    & \rotatebox{90}{\textcolor{straffic-sign}{$\blacksquare$} traf.-sign}
    & \rotatebox{90}{\textcolor{scar}{$\blacksquare$}  car}
    & \rotatebox{90}{\textcolor{struck}{$\blacksquare$}  truck}
    & \rotatebox{90}{\textcolor{sother-vehicle}{$\blacksquare$}  other-veh.}
    & \rotatebox{90}{\textcolor{svegetation}{$\blacksquare$} vegetation}
    & \rotatebox{90}{\textcolor{strunk}{$\blacksquare$}  trunk}
    & \rotatebox{90}{\textcolor{sfence}{$\blacksquare$} fence}
    & \rotatebox{90}{\textcolor{sroad}{$\blacksquare$} road}
    & \rotatebox{90}{\textcolor{ssidewalk}{$\blacksquare$} sidewalk}
    & \rotatebox{90}{\textcolor{sparking}{$\blacksquare$} parking}
    & \rotatebox{90}{\textcolor{sother-ground}{$\blacksquare$} other-grnd}
    & \rotatebox{90}{\textcolor{sbuilding}{$\blacksquare$}  building}
    & \rotatebox{90}{\textcolor{sterrain}{$\blacksquare$} terrain}
    & \rotatebox{90}{\textcolor{spole}{$\blacksquare$} pole}
    \\
\noalign{\smallskip}
\hline
\noalign{\smallskip}
JS3C-Net \cite{JS3C-Net} & C &10.31& 0.67 &0.27 & 0.00 & 0.00 & 0.00& 1.45 &24.65 & 4.41 & 6.15  &18.11&4.33 & 3.94 & 50.49& 23.74 & 11.94 & 0.07& 15.03 & 26.86 & 3.77    \\
MonoScene \cite{monoscene} & C & 11.08 & 1.86 & 1.20 & 0.00 & 0.61 & 0.45 & 2.25 & 23.26 & 6.98 & 1.48& 17.89 & 2.81 & 5.84 & 56.52 & 26.72 &  14.27& 0.46 & 14.09 & 29.64     & 4.14  \\
VoxFormer \cite{voxformer} & C & 12.35 & 1.78 & 3.32 & 0.00 & 0.59 & 0.51 & 4.18 & 25.79 & 5.63 & 3.77 & 24.39 & 5.08 & 7.64& 54.76 & 26.35 & 15.50 & 0.70 &17.65 & 29.96 & 7.11 \\ 
LMSCNet \cite{lmscnet} & L & 6.70 & 0.00& 0.00& 0.00& 0.0& 0.0& 0.00& 18.33 &0.00& 0.0& 13.66& 0.02 & 1.21&40.68& 18.22 &4.38& 0.00& 10.31 &20.54& 0.00\\
M-CONet \cite{openoccupancy}& C+L & 17.25 & 2.07 & 2.38 & 0.00 &0.66 &2.2 & 14.09 &33.25 & 19.46 & 5.99 & 32.56 &16.01 & 12.25 & 58.00 & 29.44 &19.33  & 0.02 & 27.85 &37.31 &14.88  \\
OccFusion \cite{occfusion}& C+L & 21.64& 2.33& 3.07& 0.00& 2.00 &3.39& 14.36&45.72 &19.41 & 7.92& 41.21 &19.18 & 15.74 & 65.73 & 36.46 & 21.35 &0.00 &39.32 & 46.28 & 27.60\\
OccLoff \cite{occloff} & C+L & 22.62 & 3.88 & 4.35& 0.00 &2.08&3.91& 16.47 & 46.44 & 20.38&8.72&41.20&20.06 & 15.86 & 66.25  & 43.51 & 21.07 & \textbf{0.57} & 41.23 &46.21& 27.60 \\
MS-Occ \cite{msocc} & C+L & 24.08 & 5.32 & \textbf{6.59} & 0.00 & 3.96 & 4.65 & \textbf{18.61} & 47.78 & 22.82 & 10.86 & \textbf{46.44} & 24.7 & 18.6 & 64.89  & 42.65 & 20.81 & 0.51 & \textbf{44.67} & 46.15 & 27.44 \\
\noalign{\smallskip}
\hline
\noalign{\smallskip}
TFusionOcc (T-P-25600) & C+L & 22.02 & 3.85 & 0.00 & 0.00 & 0.00 & 7.47 & 12.84 & 44.52 & 17.10 & 9.69 & 41.28 & 24.91 & 18.36 & 67.75 & 39.98 & 23.98 & 0.00 & 35.19 & 45.45 & 25.99 \\
TFusionOcc (T-SQ-25600) & C+L & 23.50 & 5.50 & 0.01 & 0.00 & 4.60 & 9.27 & 13.52 & 46.44 & \textbf{23.03} & \textbf{13.49} & 41.75 & 25.00 & 18.76 & 69.19 & 41.26 & 22.71 & 0.06 & 35.14 & 48.77 & \textbf{27.95} \\
TFusionOcc (T-SQ-IW-25600) & C+L & \textbf{24.51} & \textbf{7.93} & 0.34 & 0.00 & \textbf{10.00} & \textbf{10.19} & 13.64 & \textbf{47.82} & 16.66 & 13.43 & 42.77 & \textbf{25.84} & \textbf{19.65} & \textbf{71.19} & \textbf{43.55} & \textbf{28.21} & 0.14 & 36.89 & \textbf{49.60} & 27.79 \\
\noalign{\smallskip}
\hline
\end{tabular}}
\caption{\textbf{3D semantic scene completion performance on the SemanticKITTI validation set}. Classes are grouped to highlight performance on VRUs, Small Objects, and Complex Structures. Notation of modality: Camera (C), LiDAR (L).}
\label{occ_semanticKITTI}
\end{table*}
In both benchmarks, TFusionOcc (T-SQ-IW), despite using only a single 3D occupancy supervision signal on the current frame, achieves strong performance among object-centric approaches and remains competitive with recent state-of-the-art methods that leverage additional task supervision or temporal information, demonstrating the effectiveness and robustness of its primitive-based shape representation. Specifically, our proposed method achieves strong performance on background classes, including drivable surfaces, other flat areas, sidewalks, terrain, man-made structures, and vegetation, as well as relatively small foreground classes such as barriers, bicycles, cars, motorcycles, pedestrians, and traffic cones. This performance highlights that the proposed deformable T-primitives, particularly the T-SQ-IW ones, provide greater geometric flexibility for representing scene entities with diverse and irregular shapes. On the SurroundOcc-nuScenes benchmark, TFusionOcc outperforms DAOcc and achieves comparable performance to SDGOcc, indicating that the proposed primitive representation and multi-stage camera–LiDAR fusion can effectively exploit geometric and semantic cues without requiring auxiliary multi-task signals. On the Occ3D-nuScenes benchmark, TFusionOcc outperforms SDGOcc and remains competitive with DAOcc and ALOcc \cite{alocc}, while ALOcc \cite{alocc} additionally exploits a 16-frame temporal input. This result further suggests that the proposed geometry-aware representation and tightly coupled multi-sensor fusion provide strong occupancy modeling without relying on explicit temporal frame fusion.It is also worth noting that, as highlighted by the green and red rows in both benchmarks, under the same total number of primitives, the primitive with greater geometric flexibility achieves better performance. In addition, increasing the number of primitives from 12,800 to 25,600 generally yields further gains, although the improvement is moderate compared with that obtained from increasing primitive geometric flexibility. These results indicate that primitive shape flexibility is a more decisive factor than simply increasing primitive count in scene-level 3D semantic occupancy prediction under a fixed primitive budget.

To evaluate the applicability of TFusionOcc beyond nuScenes, we further conduct experiments on the SemanticKITTI dataset \cite{semantickitti}. As shown in Table \ref{occ_semanticKITTI}, T-SQ-IW-25600 achieves 24.51\% mIoU, outperforming the strongest camera–LiDAR baseline in our comparison, MS-Occ, by 0.43\%. Moreover, under the same 25600-primitive budget, mIoU consistently improves from 22.02\% for T-P to 23.50\% for T-SQ and 24.51\% for T-SQ-IW. This consistent trend indicates that the benefit of stronger per-primitive geometry transfers beyond the nuScenes domain and remains effective on a different driving dataset.

To further examine the performance and robustness of the proposed algorithm, we follow \cite{occfusion} to evaluate TFusionOcc in settings under challenging rainy and night-time scenarios, and the experiment results are shown in Table \ref{rainy_night}.
\begin{table}[htbp]
  \centering
  \begin{adjustbox}{width=\columnwidth}
  \begin{tabular}{c|c|c|cc}
    \toprule
    Method & Backbone & Modality & {\begin{tabular}[c]{@{}c@{}} \textbf{Rainy} \\ IoU / mIoU\end{tabular}} & {\begin{tabular}[c]{@{}c@{}} \textbf{Night} \\ IoU / mIoU\end{tabular}} \\
     \midrule 
    Co-Occ \cite{co-occ} & R101 & C+L & 40.30 / 26.60 & 35.60 / 14.60 \\
    OccFusion \cite{occfusion} & R101-DCN & C+L+R & 42.67 / 27.39 & 41.01 / 16.61 \\
    GaussianFusionOcc \cite{gaussianfusionocc} & R101-DCN & C+L & 44.28 / 29.19 & 42.78 / 18.66 \\
    GaussianFusionOcc \cite{gaussianfusionocc} & R101-DCN & C+L+R & 44.36 / 29.86 & 42.51 / 18.45 \\
    OccCylindrical \cite{occcylindrical} & Res-50 & C+L & 44.08 / 28.07 & 43.38 / 17.79 \\
    DAOcc \cite{daocc} & Res-50 & C+L & 44.51 / 29.65 & 42.85 / 18.53 \\
    \hline
    \rowcolor{green!30} TFusionOcc (T-SQ-12800) & Res-50 & C+L & 45.79 / 30.28 & 44.74 / 18.81 \\
    \rowcolor{green!30} TFusionOcc (T-SQ-IW-12800) & Res-50 & C+L & \textcolor{blue}{\textbf{45.81}} / \textcolor{blue}{\textbf{30.74}} & \textcolor{blue}{\textbf{44.82}} / \textcolor{blue}{\textbf{19.07}} \\
    \hline
    \rowcolor{red!30} TFusionOcc (T-P-25600) & Res-50 & C+L & 44.77 / 29.22 & 43.25 / 17.83 \\
    \rowcolor{red!30} TFusionOcc (T-SQ-25600) & Res-50 & C+L & 45.40 / 30.38 & 43.86 / 18.95 \\
    \rowcolor{red!30} TFusionOcc (T-SQ-IW-25600) & Res-50 & C+L & \textcolor{red}{\textbf{46.25}} / \textcolor{red}{\textbf{30.82}} & \textcolor{red}{\textbf{44.97}} / \textcolor{red}{\textbf{19.67}} \\
    \bottomrule
  \end{tabular}
  \end{adjustbox}
  \caption{\textbf{3D semantic occupancy prediction results on SurroundOcc-nuScenes validation rainy and nighttime scenario subset}. The best and second-best performance is indicated in \textcolor{red}{\textbf{bold red}} and \textcolor{blue}{\textbf{bold blue}}, respectively.}
  \label{rainy_night}
\end{table}
In both challenging scenarios, our proposed method under the T-SQ-IW setting achieved the best and second-best performance, demonstrating its robustness and the importance of the geometric shape representation capability of primitives.

\subsection{Performance Analysis on nuScenes-C Dataset}
To thoroughly assess the robustness of the proposed algorithm under extreme conditions, such as snow, fog, and sensor malfunctions, we use the nuScenes-C dataset and evaluate TFusionOcc across varying levels of corruption and severity. The results of the experiment for several camera corruption settings are shown in Table \ref{nuscenes-c-cam_IoU_mIoU}. 
\begin{table*}[htbp]
  \centering
  \footnotesize
  \begin{adjustbox}{width=\textwidth}
  \begin{tabular}{c|c|c|c|c|c|c|c|c|c}
    \toprule
    Method & {\begin{tabular}[c]{@{}c@{}} \textbf{Clean} \\ IoU / mIoU\end{tabular}} & {\begin{tabular}[c]{@{}c@{}} \textbf{Cam Crash} \\ IoU / mIoU\end{tabular}} & {\begin{tabular}[c]{@{}c@{}} \textbf{Frame Lost} \\ IoU / mIoU\end{tabular}} & {\begin{tabular}[c]{@{}c@{}} \textbf{Color Quant} \\ IoU / mIoU\end{tabular}} & {\begin{tabular}[c]{@{}c@{}} \textbf{Motion Blur} \\ IoU / mIoU\end{tabular}} & {\begin{tabular}[c]{@{}c@{}} \textbf{Bright} \\ IoU / mIoU\end{tabular}} & {\begin{tabular}[c]{@{}c@{}} \textbf{Low Light} \\ IoU / mIoU\end{tabular}} & {\begin{tabular}[c]{@{}c@{}} \textbf{Fog} \\ IoU / mIoU\end{tabular}} & {\begin{tabular}[c]{@{}c@{}} \textbf{Snow} \\ IoU / mIoU\end{tabular}} \\
     \midrule 
    OccCylindrical & 44.94 / 28.67 & 41.76 / 23.38 & 40.84 / 21.87 & 43.64 / 25.38 & \textcolor{blue}{\textbf{41.45}} / \textcolor{blue}{\textbf{22.57}} & 44.78 / 27.88 & \textcolor{red}{\textbf{43.48}} / \textcolor{red}{\textbf{22.17}} & 44.59 / 27.38 & 38.92 / \textcolor{blue}{\textbf{18.29}} \\
    DAOcc & 45.00 / 30.50 & \textcolor{red}{\textbf{42.75}} / 22.35 & \textcolor{red}{\textbf{42.16}} / 20.78 & \textcolor{blue}{\textbf{43.88}} / \textcolor{red}{\textbf{27.46}} & \textcolor{red}{\textbf{42.75}} / \textcolor{red}{\textbf{26.25}} & 43.72 / 26.78 & 41.25 / 21.15 & 43.92 / 26.87 & \textcolor{red}{\textbf{40.72}} / \textcolor{red}{\textbf{19.42}} \\
    \hline
    TFusionOcc (T-P-25600) & 45.49 / 30.04 & 42.59 / 22.34 & \textcolor{blue}{\textbf{41.97}} / 20.78 & 43.14 / 24.42 & 40.45 / 20.04 & 44.81 / 29.03 & 41.24 / 18.62 & 44.14 / 27.43 & \textcolor{blue}{\textbf{40.58}} / 15.62 \\
    TFusionOcc (T-SQ-25600) & \textcolor{blue}{\textbf{46.14}} / \textcolor{blue}{\textbf{30.88}} & 42.59 / \textcolor{blue}{\textbf{23.50}} & 41.93 / \textcolor{blue}{\textbf{22.04}} & 43.03 / 25.65 & 38.36 / 20.25 & \textcolor{blue}{\textbf{45.41}} / \textcolor{blue}{\textbf{29.92}} & 41.99 / 20.98 & \textcolor{blue}{\textbf{44.90}} / \textcolor{blue}{\textbf{28.85}} & 40.09 / 16.33 \\
    TFusionOcc (T-SQ-IW-25600) & \textcolor{red}{\textbf{47.01}} / \textcolor{red}{\textbf{31.47}} & \textcolor{blue}{\textbf{42.61}} / \textcolor{red}{\textbf{24.39}} & 41.77 / \textcolor{red}{\textbf{23.03}} & \textcolor{red}{\textbf{43.97}} / \textcolor{blue}{\textbf{26.14}} & 38.58 / 21.06 & \textcolor{red}{\textbf{46.31}} / \textcolor{red}{\textbf{30.58}} & \textcolor{blue}{\textbf{42.19}} / \textcolor{blue}{\textbf{21.67}} & \textcolor{red}{\textbf{45.95}} / \textcolor{red}{\textbf{29.69}} & 40.37 / 17.45 \\
    \bottomrule
  \end{tabular}
  \end{adjustbox}
  \caption{\textbf{3D semantic occupancy prediction results on nuScenes-C Camera Corruption benchmark}. The best and second-best performance is indicated in \textcolor{red}{\textbf{bold red}} and \textcolor{blue}{\textbf{bold blue}}, respectively.}
  \label{nuscenes-c-cam_IoU_mIoU}
\end{table*}
\begin{table*}[!t] 
\centering 
\footnotesize 
\setlength{\tabcolsep}{5.0pt} 
\renewcommand{\arraystretch}{1.20} 
\begin{adjustbox}{max width=0.75\textwidth} 
\begin{tabular}{@{}lccccccc@{}} 
\toprule 
\multirow{2}{*}{\textbf{Method}} & \multirow{2}{*}{ \makecell{\textbf{Clean}\\IoU / mIoU} } & \multirow{2}{*}{ \makecell{\textbf{LiDAR}\\ \textbf{Point Drop}\\ IoU / mIoU} } & \multirow{2}{*}{ \makecell{\textbf{Calibration}\\ \textbf{Error}\\ IoU / mIoU} } & \multirow{2}{*}{ \makecell{\textbf{LiDAR Spatial}\\ \textbf{Outliers (2\%)}\\ IoU / mIoU} } & \multicolumn{3}{c}{ \textbf{Camera Corruption + LiDAR Spatial Outliers (1\%)} } \\ \cmidrule(lr){6-8} & & & & & \makecell{\textbf{Low Light}\\IoU / mIoU} & \makecell{\textbf{Fog}\\IoU / mIoU} & \makecell{\textbf{Snow}\\IoU / mIoU} \\ 
\midrule 
G-P & 44.78 / 29.01 & 40.09 / 25.73 & 38.89 / 24.35 & 33.74 / 22.44 & 31.03 / 14.19 & 35.56 / 21.88 & 28.21 / 10.91 \\ T-P & \textbf{45.49} / \textbf{30.04} & \textbf{41.23} / \textbf{27.02} & \textbf{39.88} / \textbf{25.46} & \textbf{35.95} / \textbf{24.19} & \textbf{32.37} / \textbf{14.80} & \textbf{37.26} / \textbf{23.37} & \textbf{29.52} / \textbf{11.27} \\ 
\midrule 
G-SQ & 45.59 / 29.40 & \textbf{45.47} / 29.33 & \textbf{43.99} / 27.84 & 36.75 / 24.91 & 35.47 / 17.76 & 39.06 / 24.66 & 32.06 / 11.81 \\ T-SQ & \textbf{46.14} / \textbf{30.88} & 45.17 / \textbf{30.22} & 43.74 / \textbf{28.69} & \textbf{37.98} / \textbf{26.31} & \textbf{35.88} / \textbf{18.21} & \textbf{40.14} / \textbf{26.05} & \textbf{32.74} / \textbf{12.04} \\ \midrule 
G-SQ-IW & 46.26 / 29.98 & \textbf{46.13} / 29.89 & \textbf{44.69} / 28.39 & 37.68 / 25.49 & 36.21 / 17.72 & 40.08 / 25.27 & 33.26 / 11.44 \\ T-SQ-IW & \textbf{47.01} / \textbf{31.47} & 46.05 / \textbf{30.69} & 44.53 / \textbf{29.10} & \textbf{38.97} / \textbf{26.83} & \textbf{36.88} / \textbf{18.29} & \textbf{41.18} / \textbf{26.57} & \textbf{33.65} / \textbf{11.74} \\ 
\bottomrule 
\end{tabular} 
\end{adjustbox} 
\caption{\textbf{3D semantic occupancy prediction results on LiDAR-only and Camera--LiDAR joint corruptions benchmark.} All methods use 25,600 queries. Within each matched G/T pair, the better result for each metric is highlighted in \textbf{bold}.} 
\label{nuscenes-c-joint_IoU_mIoU} 
\end{table*}
In the camera corruption scenario, our model shows strong robustness against camera crashes, frame lost, brightness, and fog, as measured by the IoU and mIoU metric. Nevertheless, the motion blur and snow scenarios cause significant performance degradation in our model. With around 5.04\% to 8.43\% and 4.91\% to 6.64\% degradation under the IoU metric, 10.00\% to 10.63\% and 14.02\% to 14.55\% degradation under the mIoU metric for the motion blur and snow corruption case, respectively. This phenomenon could stem from the multi-stage feature fusion design strategy. Since frame lost and cam crash, both corruption results in the complete loss of information from the camera side. Our feature fusion module at each stage can easily filter out unnecessary visual features and rely solely on LiDAR features for occupancy prediction. As for brightness, fog, and low-light corruption, these conditions primarily degrade camera-side information quality and can be compensated for on the LiDAR side through our multi-stage feature fusion design. However, motion blur corruption causes structured information loss via convolution, bringing in inaccurate visual features into the feature fusion stage, resulting in a low-quality fused feature. Similarly, snow corruption introduces additional salt-and-pepper noise into the multi-stage feature fusion pipeline, resulting in a low-quality fused feature. Hence, a temporal consistency-aware visual feature aggregation, together with corruption-aware pre-filtering, may be needed for motion-blur and snow scenarios to prevent structured visual artifacts from propagating through the multi-stage fusion pipeline. Furthermore, we consistently observe that more expressive primitives achieve stronger robustness under most corruption settings, following the trend T-SQ-IW $>$ T-SQ $>$ T-P, which further highlights the importance of geometric shape flexibility.

\begin{figure*}[!h]
     \centering
     \begin{subfigure}[]{0.32\textwidth}
         \centering
         \includegraphics[width=\textwidth]{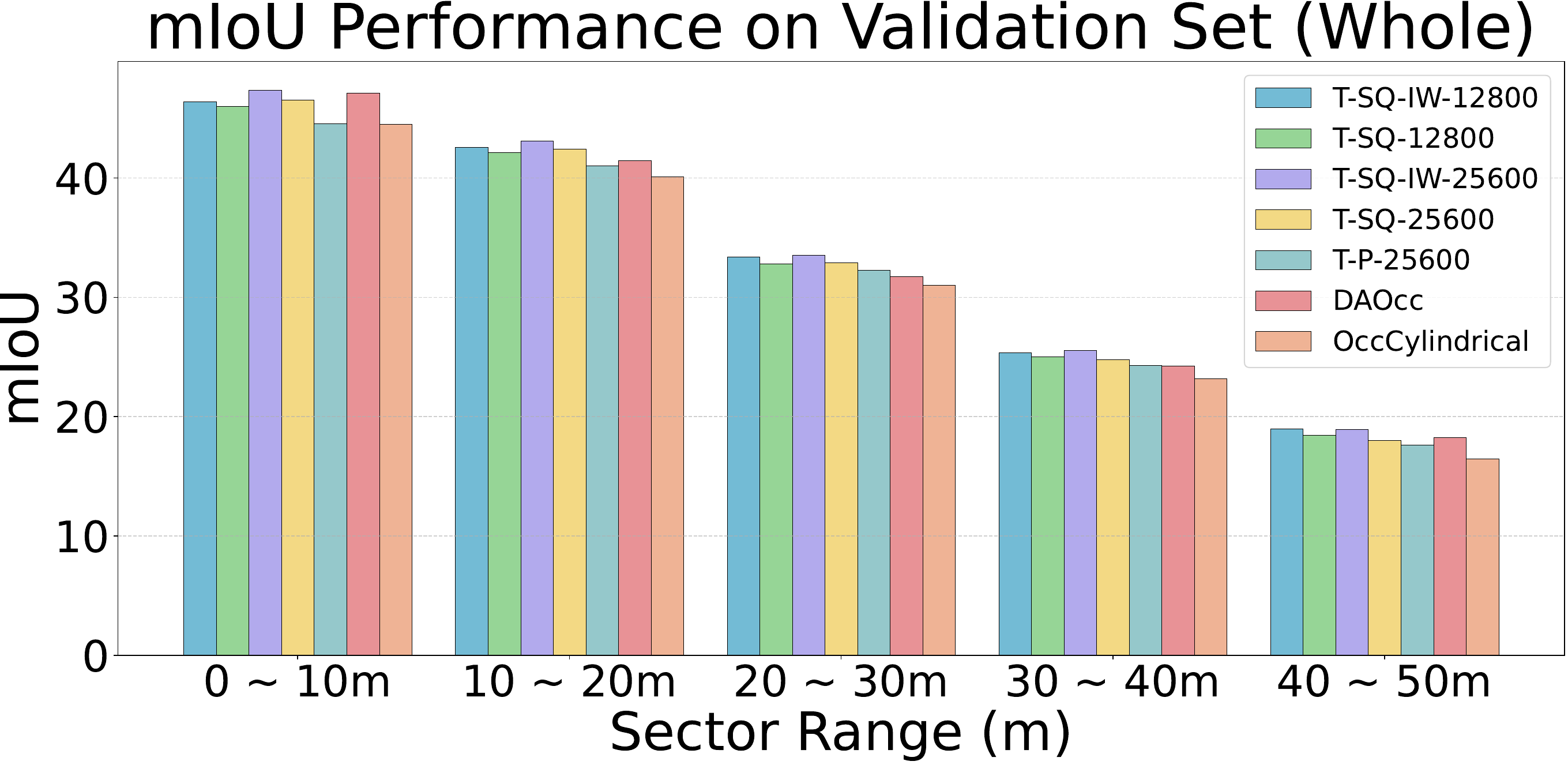}
         \caption{}
         \label{mIoU_whole_sec_trend}
     \end{subfigure}
     \begin{subfigure}[]{0.33\textwidth}
         \centering
         \includegraphics[width=\textwidth]{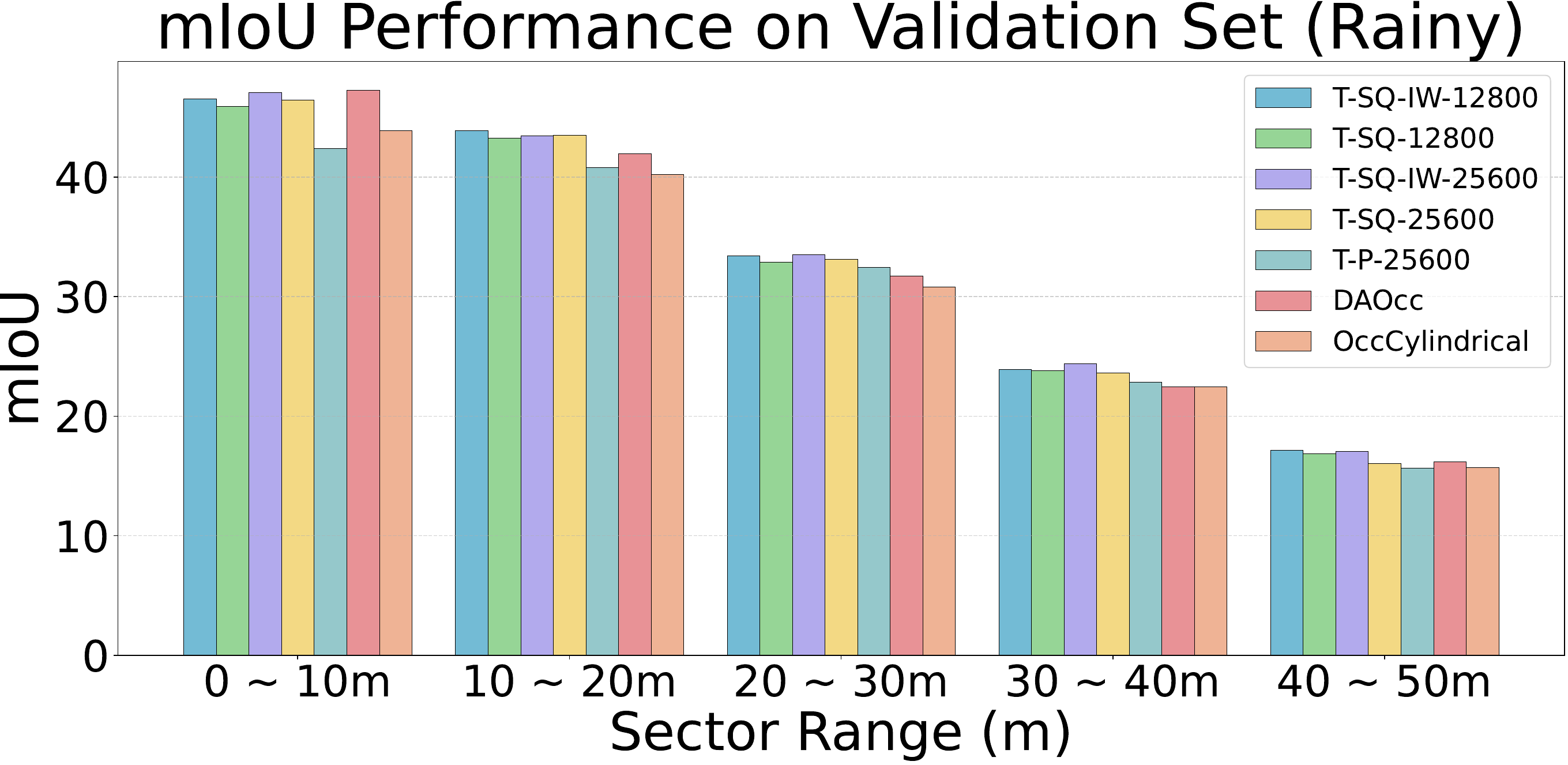}
         \caption{}
         \label{mIoU_rainy_sec_trend}
     \end{subfigure}
     \begin{subfigure}[]{0.33\textwidth}
         \centering
         \includegraphics[width=\textwidth]{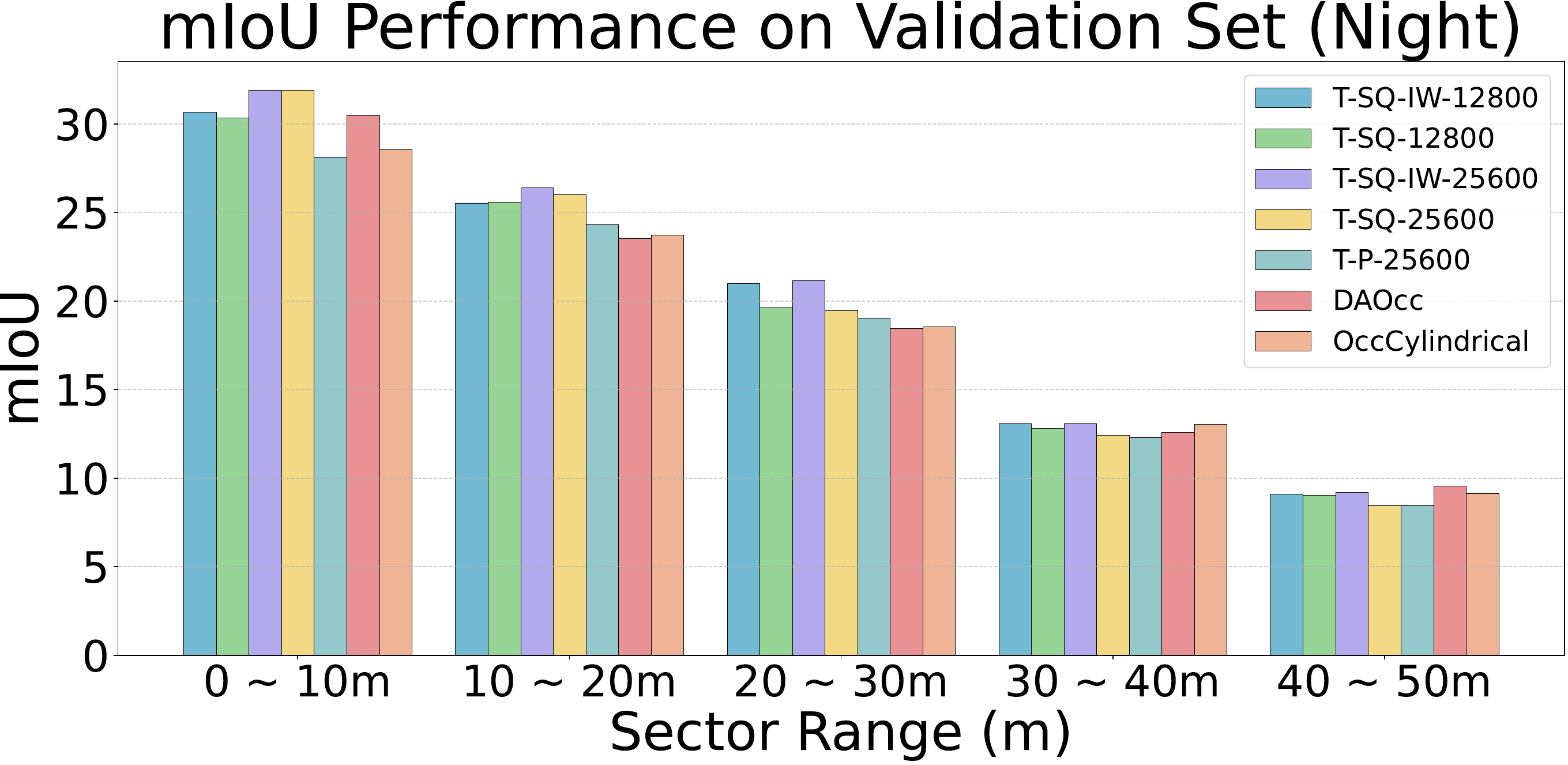}
         \caption{}
         \label{mIoU_night_sec_trend}
     \end{subfigure}
     \begin{subfigure}[]{0.32\textwidth}
         \centering
         \includegraphics[width=\textwidth]{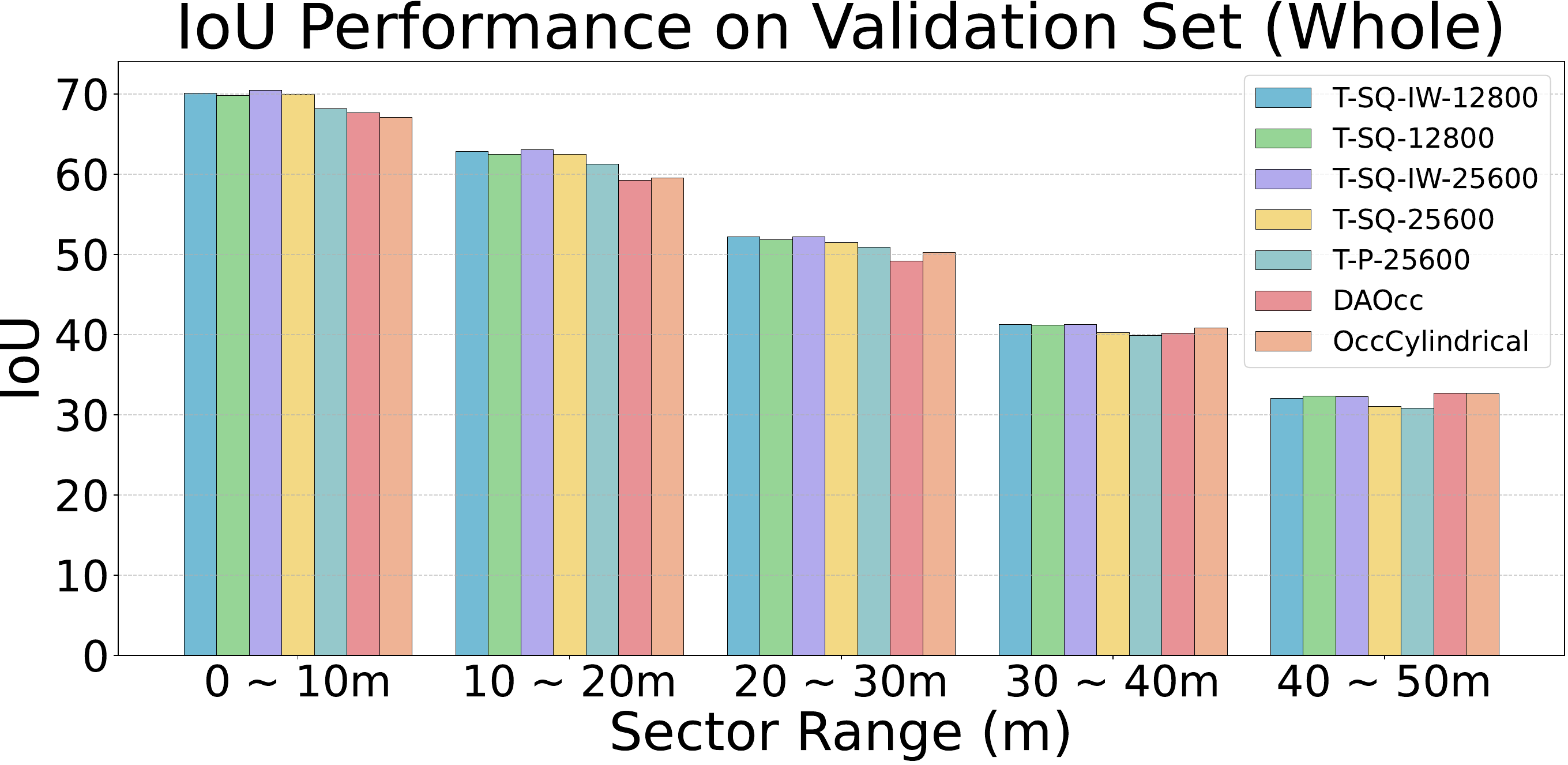}
         \caption{}
         \label{IoU_whole_sec_trend}
     \end{subfigure}
     \begin{subfigure}[]{0.33\textwidth}
         \centering
         \includegraphics[width=\textwidth]{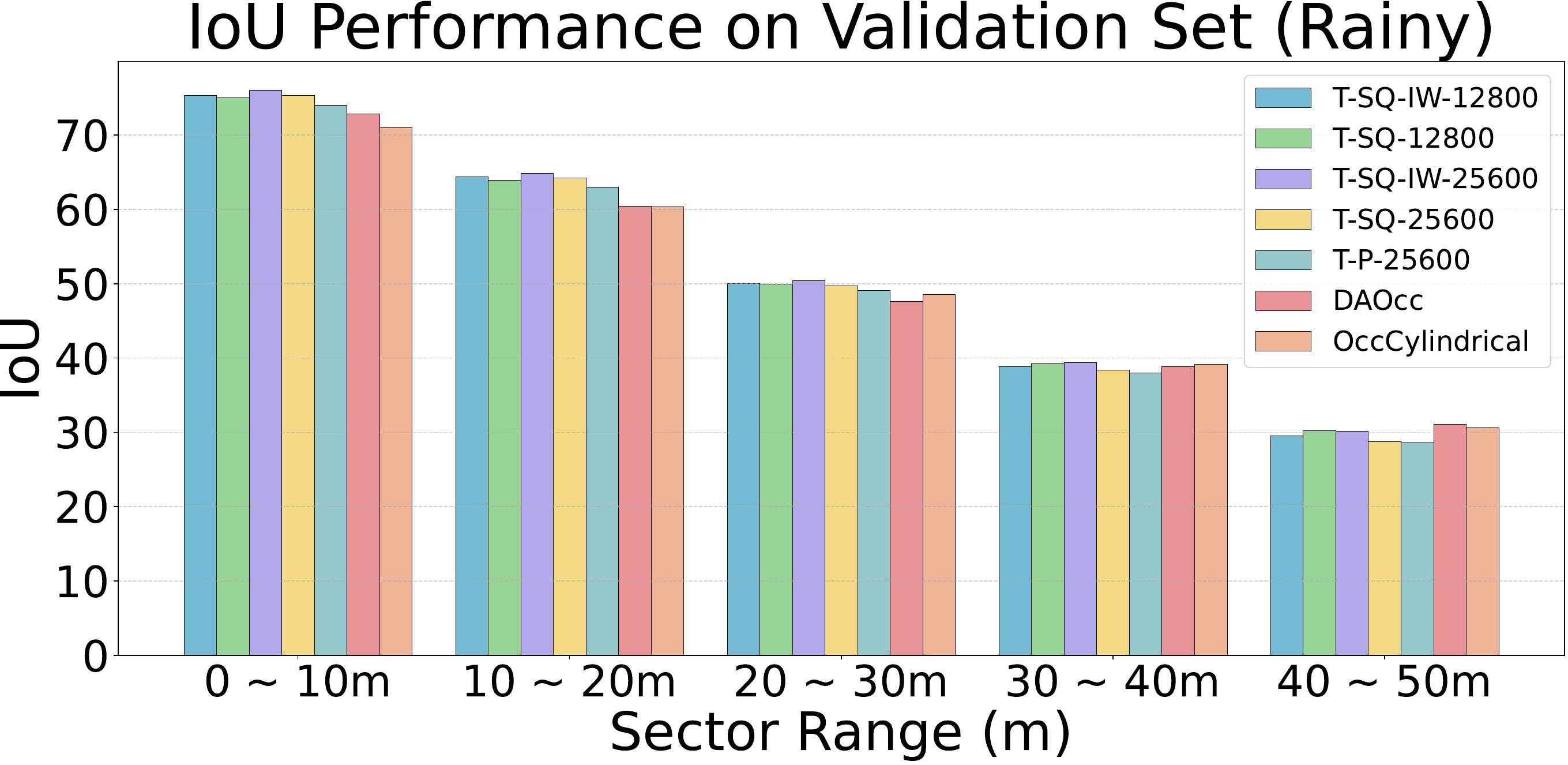}
         \caption{}
         \label{IoU_rainy_sec_trend}
     \end{subfigure}
     \begin{subfigure}[]{0.33\textwidth}
        \centering
        \includegraphics[width=\textwidth]{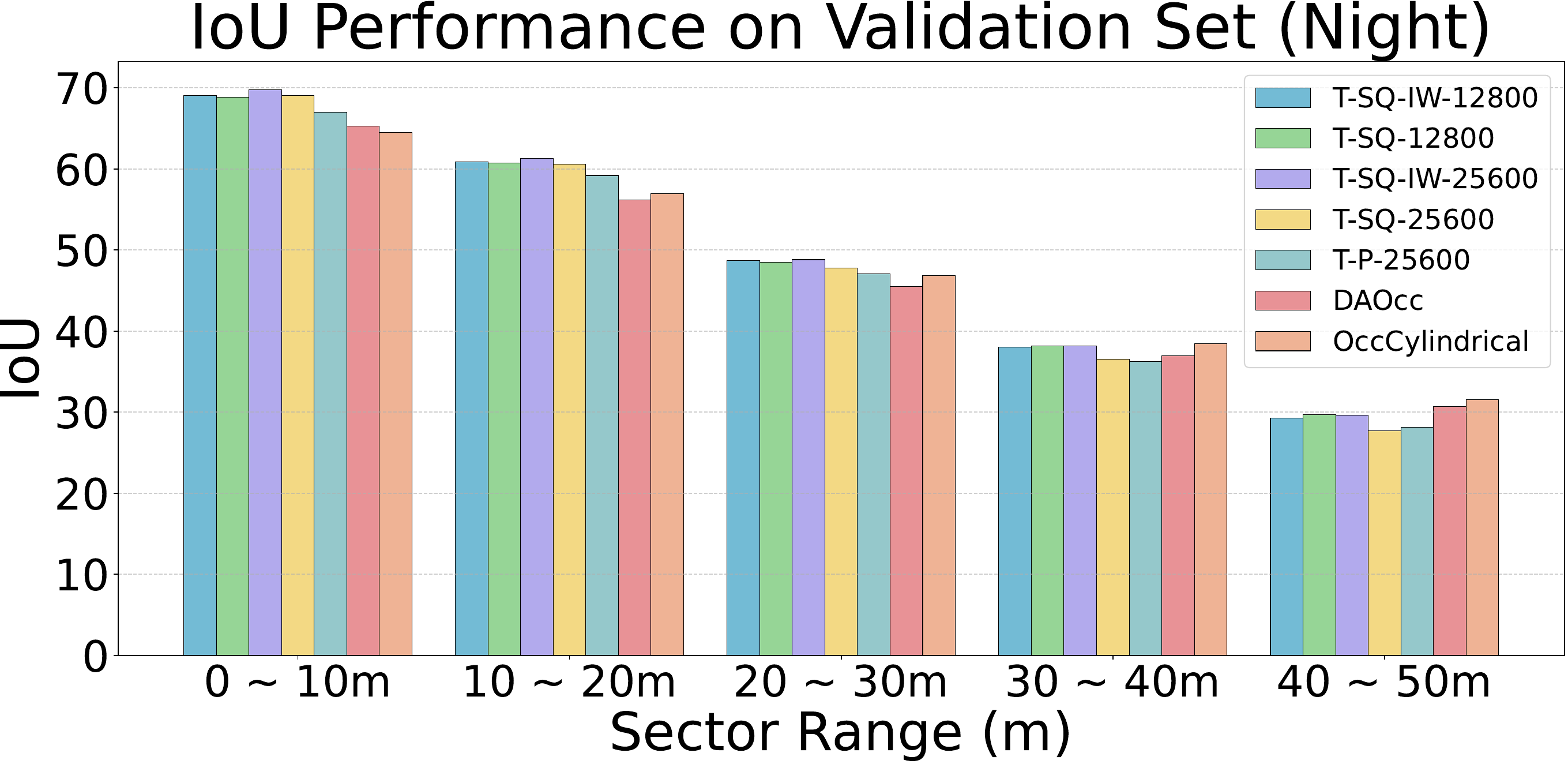}
        \caption{}
        \label{IoU_night_sec_trend}
     \end{subfigure}
        \caption{\small Performance at different sector ranges for the 3D semantic occupancy prediction task. (a) mIoU performance at different sector ranges on the whole SurroundOcc-nuScenes validation set, (b) mIoU performance at different sector ranges on the SurroundOcc-nuScenes validation rainy scenario subset, (c) mIoU performance at different sector ranges on the SurroundOcc-nuScenes validation night scenario subset, (d) IoU performance at different sector ranges on the whole SurroundOcc-nuScenes validation set, (e) IoU performance at different sector ranges on the SurroundOcc-nuScenes validation rainy scenario subset, and (f) IoU performance at different sector ranges on the SurroundOcc-nuScenes validation night scenario subset. \textbf{Better viewed when zoomed in.}}
        \label{perform_sec}
\end{figure*}

To further isolate the robustness contribution of the Student’s-t kernel, we conduct controlled Gaussian/T-kernel comparisons under LiDAR and joint camera–LiDAR corruptions, as shown in Table \ref{nuscenes-c-joint_IoU_mIoU}. For each matched G/T pair, the TFusionOcc architecture, primitive geometry, training settings, and 25,600-query budget are kept unchanged, while only the Gaussian or Student’s-t kernel is varied. We evaluate LiDAR point dropout, calibration error, 2\% LiDAR spatial outliers, and joint camera corruption with 1\% LiDAR spatial outliers. Across all tested corruptions, the T variants consistently achieve higher mIoU than their matched Gaussian counterparts. In particular, in the presence of spatial outliers, T-P, T-SQ, and T-SQ-IW improve mIoU by 1.75\%, 1.40\%, and 1.34\%, respectively. T-SQ-IW also improves mIoU by 0.80\% and 0.71\% under point dropout and calibration error, and by 0.57\%, 1.30\%, and 0.30\% under joint low-light, fog, and snow corruption, respectively. However, the gains in geometric IoU are less consistent: T-SQ and T-SQ-IW slightly underperform their matched Gaussian counterparts under LiDAR point dropout and calibration error, despite retaining higher semantic mIoU. These results indicate that the Student’s-t kernel provides a modest but consistent semantic robustness benefit, complementing the robustness contributed by primitive geometry and multi-sensor fusion.

\subsection{Performance Analysis on Varying Distance}
To thoroughly study the effect of distance on the proposed method, we evaluate our TFusionOcc against other SOTA algorithms in different sector ranges [0$\sim$10m, 10$\sim$20m, 20$\sim$30m, 30$\sim$40m, 40$\sim$50m] and present the evaluation results in Fig.~\ref{perform_sec}.
In the whole validation set, the T-SQ-IW-25600 setting dominates performance, followed by the T-SQ-IW-12800 setting under the mIoU metric, as shown in Fig.~\ref{mIoU_whole_sec_trend}. However, once the perception sector range reaches 30$\sim$40 m or 40$\sim$50 m, the geometric prediction capability under the settings T-SQ-IW-25600 and T-SQ-IW-12800 is outperformed by DAOcc and OccCylindrical, as shown in Fig.~\ref{IoU_whole_sec_trend}. In the rainy scenario, our T-SQ-IW-25600 and T-SQ-IW-12800 achieved par performance with DAOcc in the 0$\sim$10m sector range, but in the remaining sector range, the vast majority of our settings outperform DAOcc and OccCylindrical in mIoU metrics as presented in Fig.~\ref{mIoU_rainy_sec_trend}. Regarding the geometry prediction capability, our TFusionOcc, in all settings, outperforms DAOcc and OccCylindrical for the sector ranges 0$\sim$10m, 10$\sim$20m and 20$\sim$30m. However, as the perception sector range reaches 30$\sim$40 m and 40$\sim$50 m, our proposed approach is slightly left behind other SOTA algorithms, as shown in Fig.~\ref{IoU_rainy_sec_trend}. In the nighttime scenario, under the mIoU criterion, since the perception sector range is very close, like 0$\sim$10m and 10$\sim$20m, our T-SQ-IW-25600 and T-SQ-25600 achieved the best and second-best performance. As the perception sector range reaches 30$\sim$40m and 40$\sim$50m, our TFusionOcc in all settings achieved par performance with DAOcc and OccCylindrical, as shown in Fig.~\ref{mIoU_night_sec_trend}. Regarding the geometry prediction capability measured under the IoU criterion, our approach in all settings outperforms the others in the sector ranges 0$\sim$10m, 10$\sim$20m and 20$\sim$30m. Nevertheless, our approach lags behind other SOTA algorithms when the perception sector range reaches 30$\sim$40m and 40$\sim$50m as shown in Fig.~\ref{IoU_night_sec_trend}. We attribute the overall performance degradation at 30$\sim$50m mainly to reduced primitive anchor coverage and feature quality. As LiDAR observations become increasingly sparse with distance, the LiDAR-dominated Skeleton Merge initialization provides fewer reliable structural anchors in the 30$\sim$50m range, which can also limit the effective supplementation of camera anchors. Furthermore, since each primitive is evaluated within a finite rasterization region, its heavy-tailed response cannot fully compensate for insufficient or poorly localized long-range anchors.

\subsection{Qualitative Study}
\begin{figure}[t]
\centering
\begin{subfigure}[]{\columnwidth}
\includegraphics[width=\columnwidth]{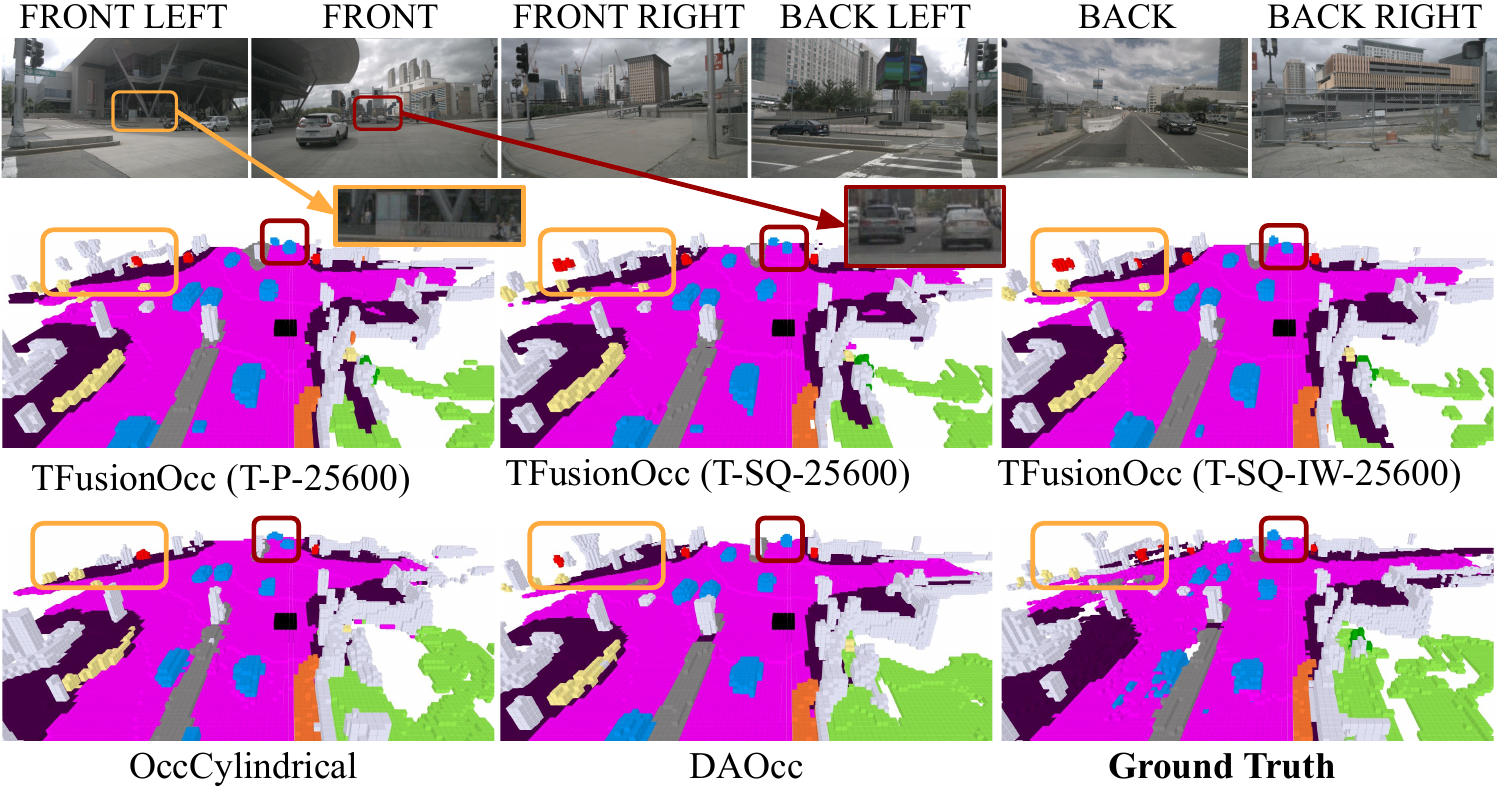}\\
\includegraphics[width=\columnwidth]{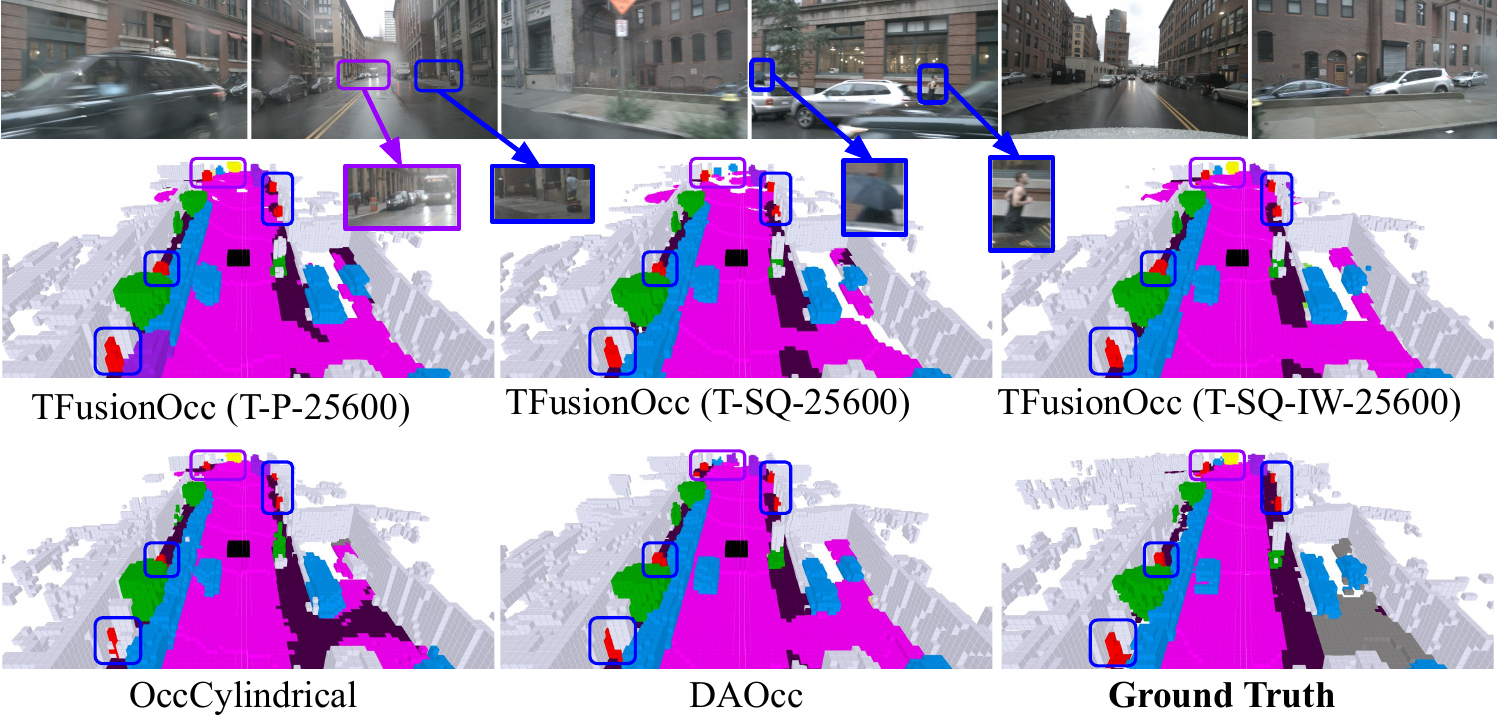}\\
\includegraphics[width=\columnwidth]{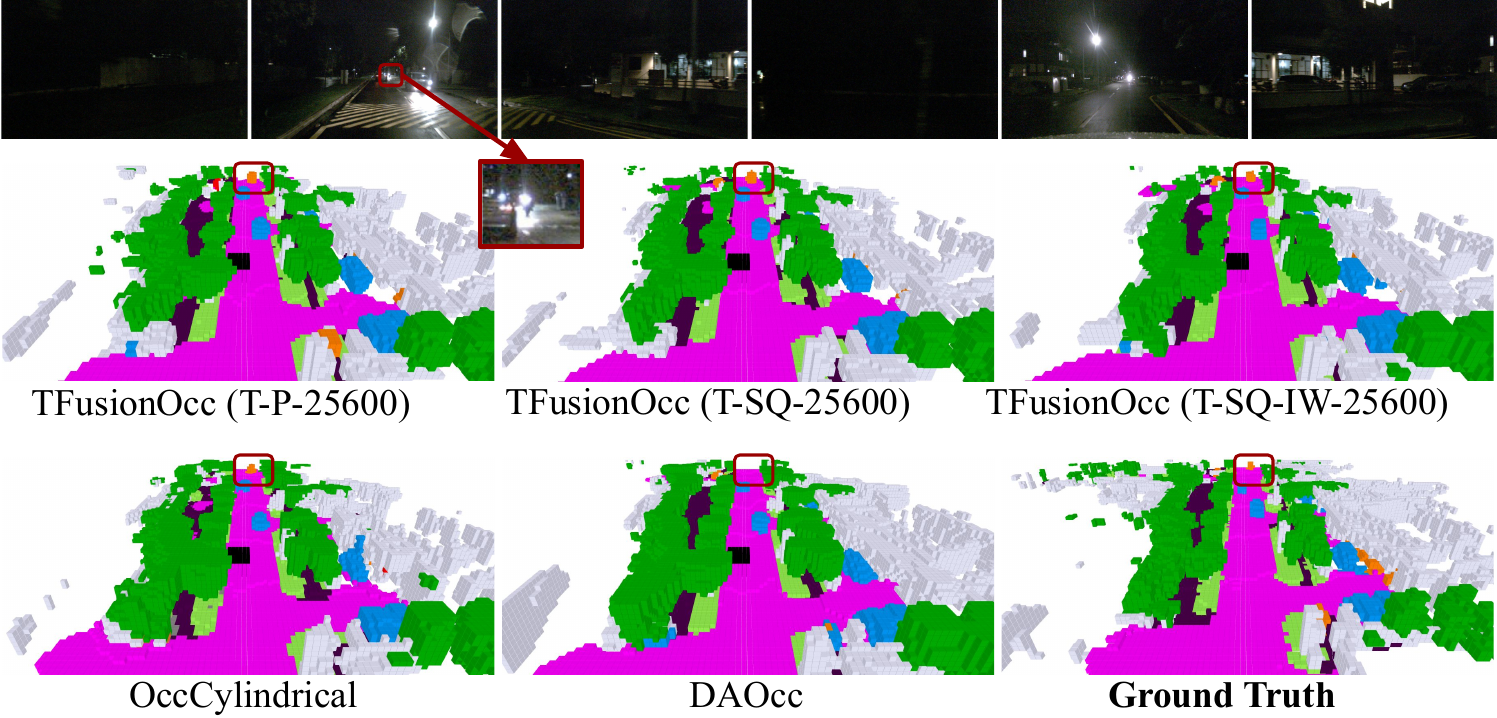}
\end{subfigure}
\caption{Qualitative results compare against other SOTA algorithms for daytime, rainy, and nighttime scenarios displayed in the upper, middle, and bottom sections, respectively. \textbf{Better viewed when zoomed in.}}
\label{compare-qua}
\end{figure}

To demonstrate the effectiveness and robustness of our method, we compare TFusionOcc in three different T-Primitive settings, such as T-P-25600, T-SQ-25600 and T-SQ-IW-25600, against other SOTA algorithms in challenging daytime, rainy and nighttime scenarios, as shown in Fig.~\ref{compare-qua}, with the primary difference area highlighted in each scenario. 
In the daytime scenario, as shown at the top of Fig.~\ref{compare-qua}, in the orange square highlighted area, benefit from multi-task learning, which provides an additional supervision signal, DAOcc successfully detects distant pedestrians and achieves similar performance as TFusionOcc. However, in the dark-red-square-highlighted area, DAOcc fails to detect the distant vehicle, demonstrating the superior performance of our proposed method. In the rainy scenario, as shown in the middle of Fig.~\ref{compare-qua}, in the purple square highlighted area, our proposed method not only successfully detects all dynamic objects, but also preserves their general contours well. In the highlighted dark-blue-square area, our approach achieves par performance with other SOTA algorithms for detecting challenging pedestrians. In the nighttime scenario, as shown at the bottom of Fig.~\ref{compare-qua}, our proposed method successfully detects very distant dynamic objects in the highlighted dark-red-square area, demonstrating its superior long-range dynamic object perception capability.

\subsection{Model Efficiency}
\begin{table}[!htpb]
  \centering
  \begin{adjustbox}{width=\columnwidth}
  \begin{tabular}{c|c|ccc}
    \toprule
    {Method} & {Modality} & {\begin{tabular}[c]{@{}c@{}}Latency\\ (ms) ($\downarrow$)\end{tabular}} & {  \begin{tabular}[c]{@{}c@{}}Memory\\ (GB) ($\downarrow$)\end{tabular}} & Params \\
    \midrule
    GaussianFormer \cite{gaussianformer} & C & 372 & 6.08 & 57.19M \\
    GaussianFormer2-256 \cite{gaussianformer2} & C & 357 & 4.43 & 125.16M \\
    \hline
    M-CONet \cite{openoccupancy} & C+L & 416.00 & 12.10 & 125.43M \\
    OccCylindrical \cite{occcylindrical} & C+L & 361.39 & 5.97 & 111.62M \\
    OccMamba-384 \cite{occmamba} & C+L & 448.30 & 12.1 & 111.08M \\
    OccMamba-128 \cite{occmamba} & C+L & \textbf{269.30} & 6.8 & 64.58M \\
    GaussianFormer3D \cite{gaussianformer3d} & C+L & 431.94 & 4.9 & 57.71M \\
    *GaussianFusionOcc \cite{gaussianfusionocc} & C+L & 460 & \textbf{2.61} & 79.63M \\
    \hline
    \rowcolor{green!30} TFusionOcc (T-SQ-12800) & C+L & 278.46 & 3.68 & 95.62M \\
    \rowcolor{green!30} TFusionOcc (T-SQ-IW-12800) & C+L & 281.47 & 3.88 & 95.96M \\
    \hline
    \rowcolor{red!30} TFusionOcc (T-P-25600) & C+L & 385.49 & 4.73 & 97.51M \\
    \rowcolor{red!30} TFusionOcc (T-SQ-25600) & C+L & 388.05 & 4.13 & 97.59M \\
    \rowcolor{red!30} TFusionOcc (T-SQ-IW-25600) & C+L & 391.06 & 4.04 & 98.24M \\
    \bottomrule
  \end{tabular}
  \end{adjustbox}
  \caption{\small Model efficiency comparison of different methods. The experiments are performed on a single RTX4090 using six multi-camera images and 10 LiDAR sweeps. For input image resolution, all methods adopt $1600\times900$. Method names marked with * indicate that their results were taken from the paper because the method is not open-source.}
  \label{efficiency}
\end{table}
We evaluate the inference efficiency of TFusionOcc under different T-primitive settings and compare its latency, memory consumption, and parameter count with representative voxel and object-centric Gaussian-based approaches in Table \ref{efficiency}. The results reveal an explicit accuracy-efficiency trade-off. Increasing the T-primitive count generally improves performance but introduces additional computational cost. Among our configurations, T-SQ-IW-12800 provides the most favorable balance among prediction accuracy, memory consumption, and inference latency, whereas T-SQ-IW-25600 achieves higher performance at the expense of increased latency. Compared with Gaussian-based object-centric approaches, TFusionOcc remains competitive in terms of resource efficiency while delivering stronger performance.

\subsection{Ablation Study}
All ablation studies use the same full TFusionOcc configuration as the reference architecture. For the kernel and primitive comparison in Table IX, the full-model baseline corresponds to the benchmark checkpoint reported in Table \ref{surroundocc-nuscenes}. The remaining ablation studies use fixed reference checkpoints of the same full-model configuration, with the corresponding baseline results reported in their respective tables. Within each study, experimental settings are kept consistent except for the component being investigated, and performance changes are measured relative to that study’s full-model baseline.
\subsubsection{Ablation Study on Different Distribution and Primitives}
\begin{table}[htpb]
    \centering
    \footnotesize
    \setlength{\tabcolsep}{4.2pt}
    \renewcommand{\arraystretch}{1.18}

    \begin{adjustbox}{max width=\columnwidth}
    \begin{tabular}{@{}l c cc cc@{}}
        \toprule
        \multirow{2}{*}{\textbf{Primitive Pair}}
        &
        \multirow{2}{*}{\textbf{Queries}}
        &
        \multicolumn{2}{c}{\textbf{IoU} ($\uparrow$)}
        &
        \multicolumn{2}{c}{\textbf{mIoU} ($\uparrow$)}
        \\
        \cmidrule(lr){3-4}
        \cmidrule(lr){5-6}

        &
        &
        \makecell{\textbf{G / T}\\ \textbf{Score}}
        &
        \makecell{\textbf{Geom. Gain}\\ \textbf{(G / T)}}
        &
        \makecell{\textbf{G / T}\\ \textbf{Score}}
        &
        \makecell{\textbf{Sem. Gain}\\ \textbf{(G / T)}}
        \\
        \midrule

        G-P / T-P
        & 25,600
        & 44.78 / \textbf{45.49}
          \;\textcolor{darkgreen}{\textbf{(+0.71)}}
        & -- / --
        & 29.01 / \textbf{30.04}
          \;\textcolor{darkgreen}{\textbf{(+1.03)}}
        & -- / --
        \\[1pt]

        G-SQ / T-SQ
        & 25,600
        & 45.59 / \textbf{46.14}
          \;\textcolor{darkgreen}{\textbf{(+0.55)}}
        & \textcolor{darkgreen}{\textbf{+0.81 / +0.65}}
        & 29.40 / \textbf{30.88}
          \;\textcolor{darkgreen}{\textbf{(+1.48)}}
        & \textcolor{darkgreen}{\textbf{+0.39 / +0.84}}
        \\[1pt]

        G-SQ-IW / T-SQ-IW
        & 25,600
        & 46.26 / \textbf{47.01}
          \;\textcolor{darkgreen}{\textbf{(+0.75)}}
        & \textcolor{darkgreen}{\textbf{+0.67 / +0.87}}
        & 29.98 / \textbf{31.47}
          \;\textcolor{darkgreen}{\textbf{(+1.49)}}
        & \textcolor{darkgreen}{\textbf{+0.58 / +0.59}}
        \\

        \bottomrule
    \end{tabular}
    \end{adjustbox}
    \caption{\small
    Controlled ablation of probability kernels and primitive geometry.
    G and T denote the Gaussian and Student's-$t$ kernels, respectively.
    Values in parentheses denote the absolute performance improvement from G to T under the same primitive geometry.
    \textit{Geom. Gain} and \textit{Sem. Gain} denotes the vertical IoU and mIoU improvements, respectively, within the G-series/T-series as primitive geometric flexibility increases.}
    \label{abla_dist}
\end{table}
Table \ref{abla_dist} presents a controlled ablation of the primitive probability kernel and geometric representation. For all variants, we keep the primitive initialization strategy, depth branch, transformer and attention modules, GCWSFusion module, loss functions, training settings, and the total number of primitive queries fixed at 25,600; only the primitive representation and its associated splatting formulation are changed. Under the same primitive geometry and primitive budget, replacing the Gaussian kernel with the corresponding Student's-$t$ kernel consistently improves performance. Specifically, the IoU gains are 0.71\%, 0.55\%, and 0.75\% for P, SQ, and SQ-IW, respectively, while the corresponding mIoU gains reach 1.03\%, 1.48\%, and 1.49\%. This indicates that the Student's-$t$ kernel provides a consistent benefit, with a more pronounced improvement in semantic occupancy prediction. Meanwhile, the vertical comparisons within both the G-series and T-series demonstrate the complementary benefit of increasing primitive geometric flexibility. From P to SQ, the IoU improves by 0.81\% / 0.65\%  and the mIoU by 0.39\% / 0.84\% for the G-series / T-series, respectively. Further introducing inverse warping from SQ to SQ-IW yields additional IoU gains of 0.67\% / 0.87\% and mIoU gains of 0.58\% / 0.59\%. Overall, these controlled results show that the Student's-$t$ kernel and increasingly expressive primitive geometry provide complementary improvements for the target scene-level 3D semantic occupancy prediction task.

\subsubsection{Ablation Study on Basis Field of T-SQ-IW Primitive}
\begin{table}[htpb]
  \centering
  \begin{adjustbox}{width=0.65\columnwidth}
  \begin{tabular}{c|cc}
    \toprule
    Basis field & IoU ($\uparrow$) & mIoU ($\uparrow$) \\
    \midrule
   $B_{1} \sim B_{24}$ & 46.23 (\textcolor{darkgreen}{\textbf{Baseline}}) & 30.77 (\textcolor{darkgreen}{\textbf{Baseline}}) \\
   $B_{10} \sim B_{24}$ & 32.98 (\textcolor{red}{\textbf{-13.25\%}}) & 23.17 (\textcolor{red}{\textbf{-7.60\%}}) \\
   $B_{1} \sim B_{18}$ & 36.20 (\textcolor{red}{\textbf{-10.03\%}}) & 24.11 (\textcolor{red}{\textbf{-6.66\%}}) \\
   $B_{1} \sim B_{9}$  & 21.83 (\textcolor{red}{\textbf{-24.40\%}}) & 11.60 (\textcolor{red}{\textbf{-19.17\%}}) \\
   No Inverse-Warp  & 20.24 (\textcolor{red}{\textbf{-25.99\%}}) & 13.87 (\textcolor{red}{\textbf{-16.90\%}}) \\
    \bottomrule
  \end{tabular}
  \end{adjustbox}
  \caption{\small Ablation study on basis field in T-SQ-IW primitive. Each range denotes the retained basis fields defined in Table~I.
$B_{10}$--$B_{24}$ excludes the constant translation, diagonal linear
scaling, and three shear fields $B_1$--$B_9$, while retaining the
remaining shear and higher-order fields.
$B_1$--$B_{18}$ excludes the additional quadratic, radial-bulging,
and mixed-coordinate fields $B_{19}$--$B_{24}$, while retaining
twisting and bending modes.
$B_1$--$B_9$ retains only constant translation, diagonal linear
scaling, and three shear fields.
No Inverse-Warp omits the entire residual warp.} 
  
  \label{abla_basis_field}
\end{table}
Table \ref{abla_basis_field} presents the results of the ablation study for each set of basis fields in the primitive T-SQ-IW. Since $B_{19} \sim B_{24}$ provides the highest level deformation, including quadratics, radial bulge, and smooth corner deformation, losing those basis fields causes 10.03\% IoU and 6.66\% mIoU degradation. In addition, because $B_{10} \sim B_{18}$ provides the second-highest level of deformation, including full shear, twist, and bending, losing $B_{10} \sim B_{24}$ results in performance drop. Surprisingly, the performance drop after losing all basis fields is similar to the loss $B_{10} \sim B_{24}$. This result is very likely because the remaining $B_{1} \sim B_{9}$ only provide basic constant, linear, shears, and taper deformation, which leads to limited improvement in the shape variety of the primitive. This result also demonstrates the importance of the high-level deformation basis field, which significantly improved the shape variety of the primitive, resulting in better overall 3D occupancy prediction performance.

\subsubsection{Ablation Study on Skeleton Merge Module}
\begin{table}[htbp]
  \centering
  \begin{adjustbox}{width=\columnwidth}
  \begin{tabular}{cccc|cc}
    \toprule
    Main-Skele & Augment & Overlap Remove & Range Filter & IoU ($\uparrow$) & mIoU ($\uparrow$) \\
    \midrule
   \Checkmark & \Checkmark & \Checkmark & \Checkmark & 46.24 (\textcolor{darkgreen}{\textbf{Baseline}}) & 30.79 (\textcolor{darkgreen}{\textbf{Baseline}}) \\
   \Checkmark & \Checkmark & \XSolidBrush & \Checkmark & 46.20 (\textcolor{red}{\textbf{-0.04\%}}) & 30.70 (\textcolor{red}{\textbf{-0.09\%}}) \\
   \Checkmark & \Checkmark & \Checkmark & \XSolidBrush & 46.07 (\textcolor{red}{\textbf{-0.17\%}}) & 30.76 (\textcolor{red}{\textbf{-0.03\%}}) \\
   \Checkmark & \Checkmark & \XSolidBrush & \XSolidBrush & 46.08 (\textcolor{red}{\textbf{-0.16\%}}) & 30.69 (\textcolor{red}{\textbf{-0.10\%}}) \\
   \Checkmark & \XSolidBrush & \XSolidBrush & \XSolidBrush & 41.96 (\textcolor{red}{\textbf{-4.28\%}}) & 29.92 (\textcolor{red}{\textbf{-0.87\%}}) \\
   \XSolidBrush & \Checkmark & \XSolidBrush & \XSolidBrush & 29.35 (\textcolor{red}{\textbf{-16.89\%}}) & 22.31 (\textcolor{red}{\textbf{-8.48\%}}) \\
    \bottomrule
  \end{tabular}
  \end{adjustbox}
  \caption{\small Ablation study on the Skeleton Merge Module. Main-Skele: LiDAR-side main skeleton branch. Augment: Camera-side augment skeleton branch.}
  \label{abla_skelemerge}
\end{table}
\begin{table}[htpb]
  \centering
  \begin{adjustbox}{width=0.65\columnwidth}
  \begin{tabular}{c|cc}
    \toprule
    Range Filter $r$ & IoU ($\uparrow$) & mIoU ($\uparrow$) \\
    \midrule
   5 m & 46.24 (\textcolor{darkgreen}{\textbf{Baseline}}) & 30.79 (\textcolor{darkgreen}{\textbf{Baseline}}) \\
   4 m & 46.21 (\textcolor{red}{\textbf{-0.03\%}}) & 30.75 (\textcolor{red}{\textbf{-0.04\%}}) \\
   3 m & 46.09 (\textcolor{red}{\textbf{-0.15\%}}) & 30.76 (\textcolor{red}{\textbf{-0.03\%}}) \\
   2 m & 45.72 (\textcolor{red}{\textbf{-0.52\%}}) & 30.69 (\textcolor{red}{\textbf{-0.10\%}}) \\
   1 m & 44.64 (\textcolor{red}{\textbf{-1.60\%}}) & 30.32 (\textcolor{red}{\textbf{-0.47\%}}) \\
   0.5 m & 43.60 (\textcolor{red}{\textbf{-2.64\%}}) & 30.04 (\textcolor{red}{\textbf{-0.75\%}}) \\
    \bottomrule
  \end{tabular}
  \end{adjustbox}
  \caption{\small Ablation study on range filter hyperparameter $r$.}
  \label{abla_range_filter_r}
\end{table}
We conducted an ablation study for each component of the skeleton merge module and presented the results in Table \ref{abla_skelemerge}. The experiment results prove the significant impact of the main skeleton structure, which supports the fundamental 29.92\% mIoU and 41.96\% IoU performance, and the auxiliary impact of the augmented skeleton, which further boosts 0.87\% mIoU and 4.28\% IoU performance. The main skeleton, based on LiDAR, provides precise 3D geometric information about the surrounding scene, especially the front faces of objects, and the augmented skeleton, based on the camera, provides additional information to mitigate the partial occlusion. Furthermore, we studied the impact of the range filter hyperparameter $r$ on the model's overall performance and present the results in Table \ref{abla_range_filter_r}. The results show that a smaller filter range forces the augmented skeleton to get too close to the main skeleton, preventing it from overcoming the LiDAR sensor's occlusion limitation, resulting in performance degradation. 

\subsubsection{Ablation Study on Fused Depth Map Guided 3D Deformable Attention Module}
\begin{table}[htpb]
  \centering
  \begin{adjustbox}{width=\columnwidth}
  \begin{tabular}{cccc|cc}
    \toprule
    Vis-Depth Fusion & Depth-Fusion & Vis-Depth & LiDAR-Depth & IoU ($\uparrow$) & mIoU ($\uparrow$)\\
    \midrule
     \Checkmark & \Checkmark & \Checkmark & \Checkmark & 46.24 (\textcolor{darkgreen}{\textbf{Baseline}}) & 30.79 (\textcolor{darkgreen}{\textbf{Baseline}}) \\
    \Checkmark & \XSolidBrush & \Checkmark & \XSolidBrush & 46.06 (\textcolor{red}{\textbf{-0.18\%}}) & 30.57 (\textcolor{red}{\textbf{-0.22\%}}) \\
     \Checkmark & \XSolidBrush & \XSolidBrush & \Checkmark & 39.80 (\textcolor{red}{\textbf{-6.44\%}}) & 25.28 (\textcolor{red}{\textbf{-5.51\%}}) \\
    \XSolidBrush & \Checkmark & \Checkmark & \Checkmark & 44.01 (\textcolor{red}{\textbf{-2.23\%}}) & 28.98 (\textcolor{red}{\textbf{-1.81\%}}) \\
    \XSolidBrush & \XSolidBrush & \Checkmark & \XSolidBrush & 43.99 (\textcolor{red}{\textbf{-2.25\%}}) & 28.93 (\textcolor{red}{\textbf{-1.86\%}}) \\
    \XSolidBrush & \XSolidBrush & \XSolidBrush & \Checkmark & 40.33 (\textcolor{red}{\textbf{-5.91\%}}) & 24.71 (\textcolor{red}{\textbf{-6.08\%}}) \\
    \bottomrule
  \end{tabular}
  \end{adjustbox}
  \caption{\small Ablation study on Depth-Fusion-Based 3D Deformable Attention Module. Vis-Depth Fusion: add the fused depth maps back to the visual features, resulting in depth-aware visual features. Depth-Fusion: fuse the visual depth maps with the LiDAR depth maps. Vis-Depth: visual depth maps. LiDAR-Depth: LiDAR depth maps.}
  \label{abla_3dfa}
\end{table}
We conducted an ablation study for each component of the fused-depth-guided 3D deformable attention module and presented the results in Table \ref{abla_3dfa}. In practice, Vis-Depth provides dense but less accurate depth maps that are good at capturing the partial occlusion of object structure. In contrast, LiDAR-Depth provides sparse but more accurate depth maps, especially good at capturing the front faces of objects in the scene. The experiment shows that Vis-Depth maps contribute significantly to the performance, even though they are less accurate. Relying solely on sparse but accurate LiDAR-Depth maps results in a performance degradation of 6.44\% IoU and 5.51\% mIoU, respectively. Furthermore, adding fused depth maps back to the original visual features results in depth-aware visual features contributing the second-largest performance improvement. The experiment results demonstrate the importance of depth map density and the necessity of multi-source depth map fusion and vis-depth feature fusion to boost model performance.

\subsubsection{Ablation Study on GCWSFusion Module}
\begin{table}[htpb]
  \centering
  \begin{adjustbox}{width=\columnwidth}
  \begin{tabular}{ccccc|cc}
    \toprule
    Gate-atten & Weight-sum & Skip-Connect & WSFusion & GCFusion & IoU ($\uparrow$) & mIoU ($\uparrow$) \\
    \midrule
    \Checkmark & \Checkmark & \Checkmark & \Checkmark & \Checkmark & 46.24 (\textcolor{darkgreen}{\textbf{Baseline}}) & 30.79 (\textcolor{darkgreen}{\textbf{Baseline}}) \\
    
    \XSolidBrush & \Checkmark & \Checkmark & \Checkmark & \Checkmark & 45.52 (\textcolor{red}{\textbf{-0.72\%}}) & 30.68 (\textcolor{red}{\textbf{-0.11\%}}) \\
    \Checkmark & \XSolidBrush & \Checkmark & \Checkmark & \Checkmark & 41.34 (\textcolor{red}{\textbf{-4.90\%}}) & 27.12 (\textcolor{red}{\textbf{-3.67\%}}) \\
    
    \XSolidBrush & \Checkmark & \Checkmark & \Checkmark & \XSolidBrush & 38.60 (\textcolor{red}{\textbf{-7.64\%}}) & 18.18 (\textcolor{red}{\textbf{-12.61\%}}) \\
     \Checkmark & \XSolidBrush & \Checkmark & \XSolidBrush & \Checkmark & 39.39 (\textcolor{red}{\textbf{-6.85\%}}) & 18.57 (\textcolor{red}{\textbf{-12.22\%}}) \\
     \Checkmark & \Checkmark & \XSolidBrush & \Checkmark & \Checkmark & 24.20 (\textcolor{red}{\textbf{-22.04\%}}) & 19.91 (\textcolor{red}{\textbf{-10.88\%}}) \\
    \bottomrule
  \end{tabular}
  \end{adjustbox}
  \caption{\small Ablation study on the GCWSFusion Module. Skip-Connet: skip-connection structure at the last feature fusion stage. WSFusion: weighted summation fusion part. GCFusion: gated concatenation fusion part. Gate-atten: Gate attention mechanism in GCFusion module. Weight-sum: Weighted summation mechanism in the WSFusion module.}
  \label{abla_gcwsfusion}
\end{table}
We conducted an ablation study for each major component of the GCWSFusion module and present the experiment results in Table \ref{abla_gcwsfusion}.
The results of the experiment show that the weight-summation mechanism plays a relatively more important role than the gate-attention mechanism in the overall GCWSFusion module. However, the WSFusion and GCFusion components contribute almost equally to the model's final performance, with WSFusion slightly more important. Missing neach part will cause 12.22\% to 12.61\% mIoU and 6.85\% to 7.64\% IoU performance degradation. The skip-connection structure also plays an important role during feature fusion; without it, the IoU and mIoU degrade by 22.04\% and 10.88\%, respectively.

\section{Conclusion}\label{conclusion}
This paper presents TFusionOcc, a T-primitive-based object-centric multi-sensor fusion framework for 3D semantic occupancy prediction. The proposed method is built upon a family of Student’s t-distribution-based T-primitives, among which the deformable T-Superquadric with inverse warping serves as the key geometry-enhancing primitive. To further improve robustness, we introduce a unified probabilistic formulation based on Student’s t-distribution and the T-mixture model for joint occupancy and semantic modeling, together with a tightly coupled multi-stage fusion architecture for effective camera–LiDAR integration. Extensive experiments on SurroundOcc-nuScenes and Occ3D-nuScenes demonstrate competitive occupancy prediction performance, while additional evaluation on SemanticKITTI further validates the applicability of TFusionOcc across different datasets and sensor configurations. In particular, the consistent improvement from T-P to T-SQ and T-SQ-IW across both nuScenes and SemanticKITTI-based evaluations confirms the task-level benefit of stronger primitive geometry. Experiments on nuScenes-C further demonstrate robustness under a range of challenging corruption scenarios. Future work will investigate temporal consistency-aware visual feature aggregation and corruption-specific visual preprocessing to suppress structured artifacts caused by motion blur and snow prior to multi-sensor fusion.

\bibliographystyle{IEEEtran}
\bibliography{TFusionOcc-refs}

\begin{IEEEbiography}[{\includegraphics[width=1in,height=1.25in, clip,keepaspectratio]{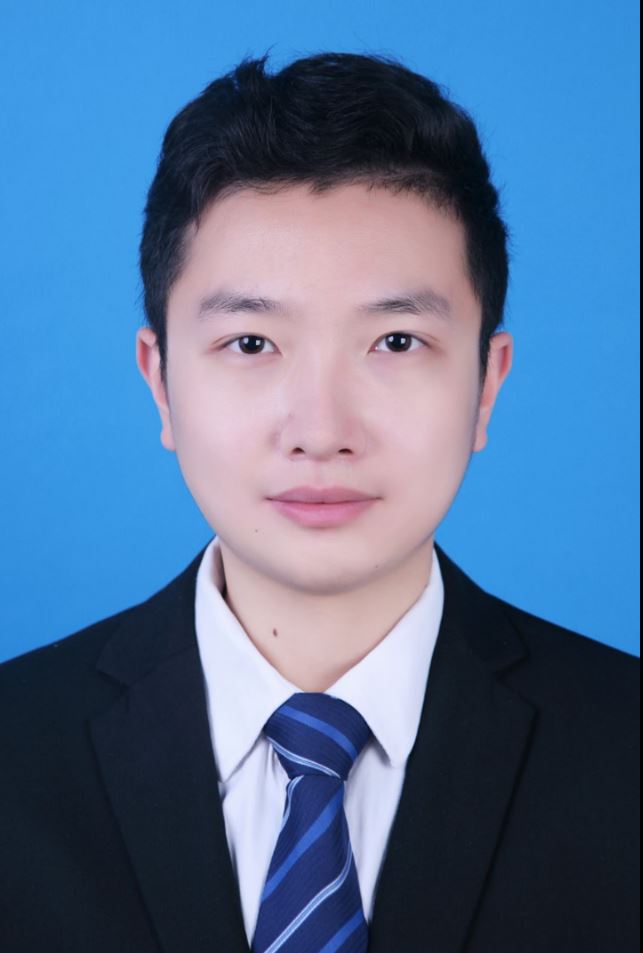}}]{Zhenxing Ming}
Zhenxing Ming received the B.Sc. degree in electrical engineering from the Hubei University of Technology, Wuhan, China, in 2018, and the master’s degree from the University of New South Wales, Sydney, NSW, Australia, in 2021. He is currently working toward the Ph.D. degree with the Australia Centre for Robotics (ACFR), The University of Sydney, Sydney, NSW. He was an AV Engineer with Li Auto Inc., Beijing, China. His research focuses on multimodality-based environment perception and sensor fusion for AV.
\end{IEEEbiography}

\begin{IEEEbiography}[{\includegraphics[width=1in,height=1.25in, clip,keepaspectratio]{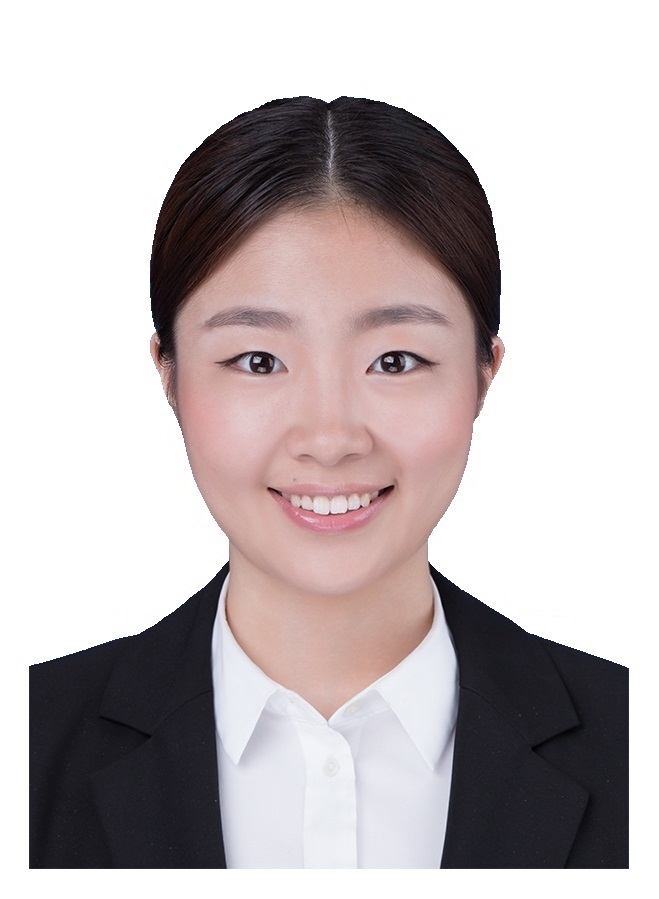}}]{Yaoqi Huang} received the B.S. degree from Fudan University, Shanghai, China, in 2016, and the M.S. degree from The University of Melbourne, Melbourne, VIC, Australia, in 2018. She is currently pursuing the Ph.D. degree with the Australia Centre for Robotics (ACFR), The University of Sydney, Sydney, NSW. Her research interests lie in traffic scene understanding, with an emphasis on developing methodologies to analyze and 
interpret attention-worthy traffic entities for intelligent driving systems. 
\end{IEEEbiography}

\begin{IEEEbiography}[{\includegraphics[width=1in,height=1.25in, clip,keepaspectratio]{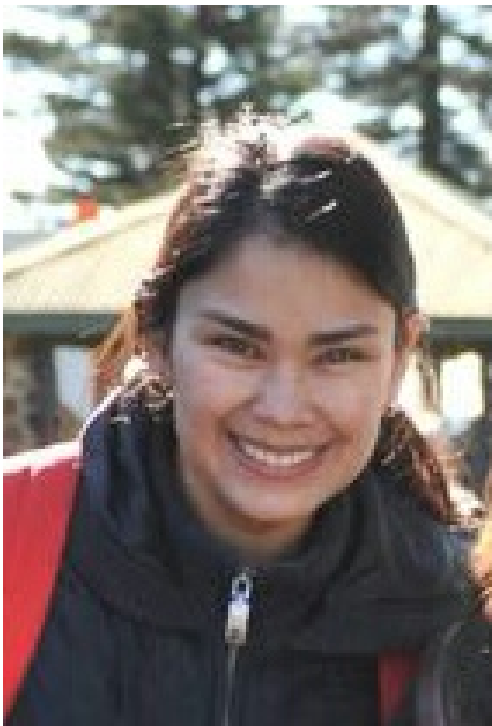}}]{Julie Stephany Berrio} received the B.S. degree in Mechatronics Engineering in 2009 from Universidad Autonoma de Occidente, Cali, Colombia, and the M.E. degree in 2012 from the Universidad del Valle, Cali, Colombia. She received her PhD from the University of Sydney, Australia, in 2021. She is a Senior Lecturer with the School of Mechanical and Mining at the University of Queensland. Her research interests include perception and mapping for intelligent transportation systems.
\end{IEEEbiography}

\begin{IEEEbiography}[{\includegraphics[width=1in,height=1.25in,clip,keepaspectratio]{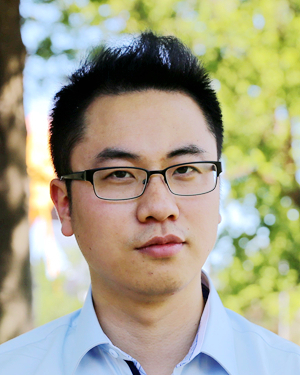}}]{Mao Shan} received the B.S. degree in electrical engineering from the Shaanxi University of Science and Technology, Xi’an, China, in 2006, and the M.S. degree in automation and manufacturing systems and PhD degree from the University of Sydney, Australia, in 2009 and 2014, respectively. He is a Research Fellow with the Australian Centre for Robotics at the University of Sydney. His research interests include autonomous systems, V2X, and tracking algorithms and applications.
\end{IEEEbiography}

\begin{IEEEbiography}[{\includegraphics[width=1in,height=1.25in,clip,keepaspectratio]{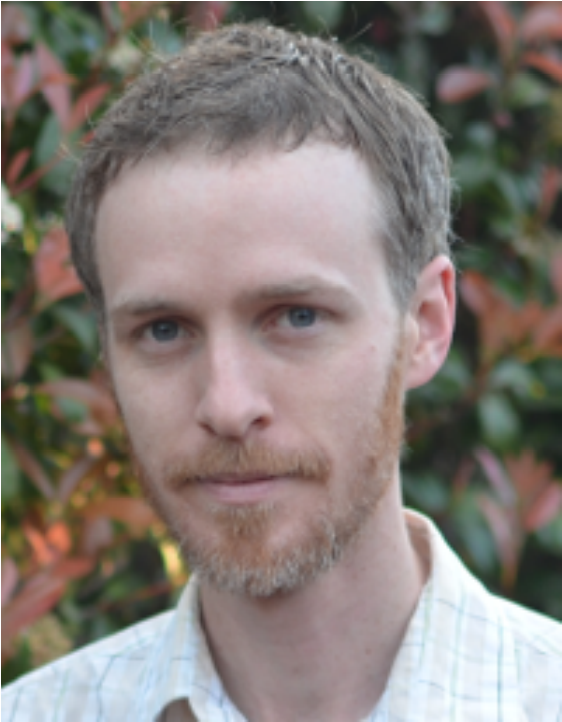}}]{Stewart Worrall} received the PhD from the University of Sydney, Australia, in 2009. He is currently a Senior Research Fellow with the Australian Centre for Robotics, University of Sydney. His research is focused on studying and applying technology for vehicle automation and improving safety.
\end{IEEEbiography}

\end{document}